\documentclass{article}

\usepackage{PRIMEarxiv}

\usepackage[utf8]{inputenc} 
\usepackage[T1]{fontenc}    
\usepackage{hyperref}       
\usepackage{url}            
\usepackage{booktabs}       
\usepackage{amsfonts}       
\usepackage{nicefrac}       
\usepackage{microtype}      
\usepackage{graphicx}
\usepackage{amsmath}
\usepackage{xcolor}
\definecolor{darkgreen}{RGB}{0,100,0} 
\usepackage{float}
\usepackage{tikz}
\usepackage{multirow}
\usepackage{tabularx}
\usetikzlibrary{positioning,chains, trees, calc, arrows}
\usepackage{tikz-qtree}
\usepackage{pdfpages}
\usepackage{natbib}
\usepackage[linesnumbered,ruled,vlined]{algorithm2e}
\usepackage{longtable}
\graphicspath{{media/}}     

\begin{document}

\title{ALICE: Combining Feature Selection and Inter-Rater Agreeability for Machine Learning Insights

}

\author{
  Bachana Anasashvili \\
  Humboldt University of Berlin \\
  \texttt{bachana.anasashvili@student.hu-berlin.de} \\
    \And
  Vahidin Jeleskovic \\
  Humboldt University of Berlin \\
  \texttt{vahidin.jeleskovic@hu-berlin.de} \\
}

\maketitle

\begin{abstract}
This paper presents a new Python library called Automated Learning for Insightful Comparison and Evaluation (\texttt{ALICE}), which merges conventional feature selection and the concept of inter-rater agreeability in a simple, user-friendly manner to seek insights into black box Machine Learning models. The framework is proposed following an overview of the key concepts of interpretability in ML. The entire architecture and intuition of the main methods of the framework are also thoroughly discussed and results from initial experiments on a customer churn predictive modeling task are presented, alongside ideas for possible avenues to explore for the future. The full source code for the framework and the experiment notebooks can be found at: \href{https://github.com/anasashb/aliceHU}{https://github.com/anasashb/aliceHU}.

\end{abstract}

\section{Introduction}
\label{sec:intro}
The use of Machine Learning models for decision-making has become the new norm not only in tech but any business field imaginable, covering any possible task at hand be it search engine recommendations, customer churn prediction, credit risk scoring, energy load forecasting, or the deployment of personalized AI assistants. This comes at a time when developing ML models has become increasingly easier with the rise of open-source, free and user-friendly Python libraries such as \texttt{Keras}, \texttt{scikit-learn}, \texttt{PyTorch} and as generative AI-based conversational chatbots such as ChatGPT, Gemini and Claude that can provide coding assistance --- if not ready-made code for modeling --- are evolving rapidly. 

Such developments yet again beg the question of interpretability in machine learning, which has been formulated in various ways in literature and been offered multiple proposed solutions such as exploring \emph{causality} (see Section~\ref{subsec:causal_ml}), \emph{explainability} (see Section~\ref{subsec:xai}) or abandoning black box ML models altogether. But to make a philosophical argument, it is hard to see the benefits of highly model or domain-specific, post-hoc, or complex solutions to obtain insights into the inner-doings of machine learning models when the modeling task itself is growing ever more accessible to laypeople.

Common thought on categorizing ML models in this regard would argue that parametric models descending from the fields of statistics and econometrics such as Linear or Logistic Regression are by nature more interpretable than their data-driven and non-parametric counterparts such as tree-based models or neural networks. While we do not in any form challenge this widely-held assumption, we inquire if there's a possibility to gauge the degree of differences between the outcomes from these different models. 

There are troves of literature that compare a vast array of learning algorithms on this or that predictive modeling task offering analysis based on error metrics solely. Yes, when deploying a model in a business, the best performance is what may matter most --- after all, the smaller the error the better. Let us assume an overly simplified scenario where an ML engineer is handed data with a sizeable amount of features and tasked to build a predictive model. We would not be incorrect to assume that in its most skeletal form, the following process would include some sort of feature selection, evaluation runs with different model architectures, and a final decision based on which model performs best on the validation set. While knowing that a hypothetical Random Forest performed better on the hypothetical task than a hypothetical Logistic Regression or a Multi-Layer Perceptron may be sufficient enough, what we suggest is that it could also be \emph{beneficial} to know, along the way, how much these models agreed with each other --- which can serve as an additional layer of insight in the overall process.

Given the aforementioned, and with user-friendliness in mind, this paper proposes Automated Learning for Insightful Comparison and Evaluation (\texttt{ALICE}), a novel framework that fuses feature selection and inter-rater agreeability in a simplified, automated manner. 

\section{Background and Related Work}
\label{sec:background_related_work}

Over the scope of the recent decades, a significant portion of research has been dedicated to probing the explainability, interpretability, or causality of Machine Learning models. Such research has diverged in several different directions which we will briefly categorize throughout this section. As regards the structure, we'll proceed with briefly reviewing and motivating causal ML in Section~\ref{subsec:causal_ml}, explainable AI in Section~\ref{subsec:xai}, and inter-rater agreeability in Section~\ref{subsec:reliable_ml}. Besides that we will motivate the concept of Feature Selection in Section~\ref{subsec:feature_selection}, as it will serve as a key foundation of the experimental framework proposed in Section~\ref{subsec:general_overview}.

\subsection{Causal ML}
\label{subsec:causal_ml}
The concept of causality in Machine Learning is deeply tied to the field of causal discovery. The main essence of the said field is to determine causal relationships between a given set of variables, which can be done by mapping the features onto a directed acyclic graph (DAG) structure where an edge pointing from node \(A\) to \(B\) means the former causes --- is the parent of --- the latter, and the latter is a child of the former. Another term is a spouse, which corresponds to hypothetical nodes \(A\) and \(C\) both point toward \(B\) but are not otherwise related. 

A key survey paper that covers in-depth how causal discovery can be used for feature selection is given in \citet{yu_causality-based_2020}. One of the main concepts mentioned by the authors in the survey is the Markov Boundary, which consists of the parents, children, and spouses of some target variable and serves as the "minimal feature subset with maximum predictivity" for the modeling task.

There are multiple algorithms developed over the years for causal discovery, which \citet{yu_causality-based_2020} classify into constraint and score-based methods. The former methods obtain the DAG relationships between the variables using statistical tests of conditional independence while the latter algorithms optimize some score functions such as the Akaike Information Criterion or the Bayesian Information Criterion instead. \citet{glymour_review_2019} lists Peter-Clark (PC) algorithm by \citet{spirtes_algorithm_1991} and Fast Causal Inference (FCI) algorithms by \citet{spirtes_anytime_2001} as key examples of constraint-based methods and the Greedy Equivalence Search (GES) algorithm by \citet{chickering_optimal_2003} as a notable example of the score-based methods.

One key challenge, of the field, however, when it comes to ML is that causal discovery is not a field that is covered well by Python libraries. If Java, R, and MATLAB all have well-renowned and established libraries --- \texttt{TETRAD}, \texttt{pcalg}, \texttt{Bayes Net Toolbox}, respectively --- this is not at all the case of Python, which makes it hard to integrate such approaches in modern ML pipelines built around the language. A promising library for such purpose is \texttt{casual-learn} announced recently in a preprint by \citet{zheng_causal-learn_2023}.

\subsection{Explainable AI}
\label{subsec:xai}
Explainable AI (XAI) is yet another research field that has gathered steam in light of the growth of the popularity of Machine Learning solutions. 
For a point of reference, two comprehensive review papers that look into the field are \citet{adadi_peeking_2018} and \citet{gilpin_explaining_2018}. \citet{adadi_peeking_2018} denotes that explainable AI methods can fall (among various possible categorizations) into either \emph{post-hoc} or \emph{intrinsic} categories, where the former means the decisions of the black box ML model are attempted to be explained after they are made, while the latter implies somehow making the modeling framework interpretable from the get-go. A popular example of \emph{post-hoc} XAI technique, which greatly contributed to the emergence of the field altogether, is Local Interpretable Model-agnostic Explanations (LIME; \citep{ribeiro_why_2016}), which employs simple, linear proxy models to approximate the decision-making of an ML model as regards a specific observation from the input dataset. Another popular \emph{post-hoc} XAI method is Layer-wise Relevance Propagation (LRP; \citep{bach_pixel-wise_2015}), which attempts to determine how inputs determined an outcome for a specific observation by tracing back layer-by-layer from the prediction. \citet{gilpin_explaining_2018} mentions the Attention Mechanism as one of the key methods for \emph{Intrinsic} explainability, which is a main component of the Transformer \citep{vaswani_attention_2017} neural network architecture and has seen tremendous success in Natural Language Processing as evidenced by, among other models, BERT \citep{devlin_bert_2019} and GPT \citep{brown_language_2020}.

\subsection{Inter-Rater Agreeability}
\label{subsec:reliable_ml}
The concept of inter-rater agreeability is deeply tied to the medical field, for which initially \citet{cohen_coefficient_1960} proposed a measure to assess agreeability between the medical assessments (categorical variables with \(2+\) classes) by two judges. Since then, the proposed coefficient, \(\kappa\) has remained somewhat of a staple in the field and has seen several variations such as intra-class \(\kappa\), weighted \(\kappa\), or competing metrics such as the tetrachoric correlation coefficient \citep{banerjee_beyond_1999}. 

In its initial form, the proposed metric by \citet{cohen_coefficient_1960} is computed as follows:

\begin{align}
    \kappa = \frac{p_o - p_e}{1 - p_e},
\end{align}

where \(p_o\) stands for the proportion of observed agreement between raters, and \(p_e\) for the proportion of chance agreement between raters. Therefore the difference \(p_o - p_e\) is defined as "the proportion of cases in which beyond-chance agreement occurred." The upper limit of the metric is \(1\), which denotes complete agreement between two systems, while \(0\) denotes no agreement besides that which can be accredited to chance. The original paper cannot define a lower limit but notes that negative values are technically possible and would imply that "observed agreement is less than expected by chance."

For some binary categorical variable \(c \in \{0, 1\}\), we can demonstrate how \(p_0\) and \(p_e\) are computed by generalizing a specific clinical example given in \citet{sim_kappa_2005}. Given two raters (in the specific example they were clinicians) and given a binary variable, let \(a\) denote the case when rater \(1\) and rater \(2\) both asses \(c=1\), \(b\) the case when judge \(1\) rates \(c=1\) and judge \(2\) rates \(c=0\), \(c\) the case when judge \(1\) rates \(c=0\) and judge \(c=1\), and \(d\) the case when judge \(1\) and \(2\) agree that \(c=0\). Let the number of observations \(n=39\) of which occurrences of the four cases are: \(a=22\), \(b=2\), \(c=4\), \(d=4\).

Then, the proportion of observed agreement between the two raters can be computed as the number of times they agreed out of total observations: 

\begin{align}
    &p_o = \frac{a+d}{n} = \frac{22+11}{39} = 0.8462.
\end{align}

Meanwhile, the proportion of chance agreement between the raters can be computed as:

\begin{align}
    & p_e = \frac{ \frac{(a + c) \times (a + b)}{n} + \frac{(b + d) \times (c + d)}{n}}{n} \\ \nonumber
    & = \frac{(a + c) \times (a + b) + (b + d) \times (c + d)}{n^2} \\ \nonumber
    & = \frac{26 \times 24 + 13 \times 15}{39^2} \\ \nonumber
    & = \frac{624 + 195}{1,521} \\ \nonumber
    & = 0.5385
\end{align}

Given the values of \(p_o\) and \(p_e\), the agreeability measure can then be computed as:

\begin{align}
    \kappa = \frac{0.8462 - 0.5385}{1 - 0.5385} = 0.67.
\end{align}

From the initial proposal as a means to evaluate agreeability between two different human raters, Cohen's \(\kappa\) has also been adapted to operate with a confusion matrix of a binary classification task in Machine Learning and used to assess the agreement between true and predicted classes from a classifier \citep{delgado_why_2019}. Some other works that also use and evaluate the fitness of \(\kappa\) as a performance metric are \citet{ben-david_about_2008, chicco_matthews_2021}. It should be noted at this point that in our experimental framework (refer to Section~\ref{subsec:general_overview}) \(\kappa\) is used with its original purpose of measuring agreement between the ratings of two classifiers and not as a metric between true and predicted classes.

The \emph{interpretation} of \(\kappa\) varies but based on \citet{viera_understanding_2005, landis_measurement_1977} it can be formulated generally as: \(\kappa\) below 0 indicates no agreement, between 0 and 0.20 indicates slight agreement, between 0.21 and 0.40 indicates fair agreement, between 0.41 and 0.60 indicates moderate agreement, between 0.61 and 0.80 indicates substantial agreement, between 0.81 and 1.00 indicates almost perfect agreement.

\subsection{Feature Selection}
\label{subsec:feature_selection}
A good categorization of conventional feature selection methods can be defined in accordance with \citet{venkatesh_review_2019} as filter and wrapper methods\footnote{Here we skip over embedded methods as they are beyond the scope of the proposed framework.}. The former are methods that select features to include in a predictive model based on statistical measurements including but not limited to information gain, \(\chi^{2}\) tests, Pearson's correlation (\(\rho\)), and Fisher score. While these techniques are not too computationally costly, wrapper methods have been perceived as more accurate with the trade-off of an increase in computational complexity. What the latter methods do is select features based on the results of a model, or specifically the error metrics. 

The use of wrapper methods involves a search, which can have two directions --- either sequential or random. The three most conventional methods of sequential direction, along with pseudo-code, are described in an early work from the field of feature selection by \citet{ferri_comparative_1994}. These are Sequential Forward Selection \citep{whitney_direct_1971}, Sequential Backward Elimination \citep{marill_effectiveness_1963}, and the Plus-\(l\)-Take-Away-\(r\) algorithm \citep{d_selecting_1976}.

While such wrapper methods, approximated, lead to \(\mathcal{O}(n^2)\) complexity, where \(n\) stands for the number of variables in a given feature set, they are aimed at overcoming the NP-hard \citet{venkatesh_review_2019} challenge posed by conducting an exhaustive, greedy search that would try out all possible \(2^n\) combinations. The issue of \(\mathcal{O}(2^n)\) complexity can be easily demonstrated: if our input feature set carries \(20\) variables, there are exactly \(2^{20}=1,048,576\) possible combinations that could be tried out in modeling.

\section{Proposed Framework}
\subsection{General Overview}
\label{subsec:general_overview}
Given the review above, the proposed framework draws inspiration from using simple models against more complex non-parametric models (Section~\ref{subsec:xai}) for obtaining insights into the black box of ML. But instead of XAI methods it employs inter-rater agreeability, which is combined at every step of feature selection. Techniques of causal feature selection (Section~\ref{subsec:causal_ml}) are not incorporated due to a lack of well-established Python frameworks. Besides, the aim of the framework is to establish a user-friendly, intuitive, and simpler approach to get further insights --- which can be achieved by using the most standard and basic feature selection methods. Currently, the framework uses Backwards Elimination, and hence in the following parts any mention of the feature selection procedure will imply that.

\begin{figure}[h] 
  \centering
  \includegraphics[width=\textwidth]{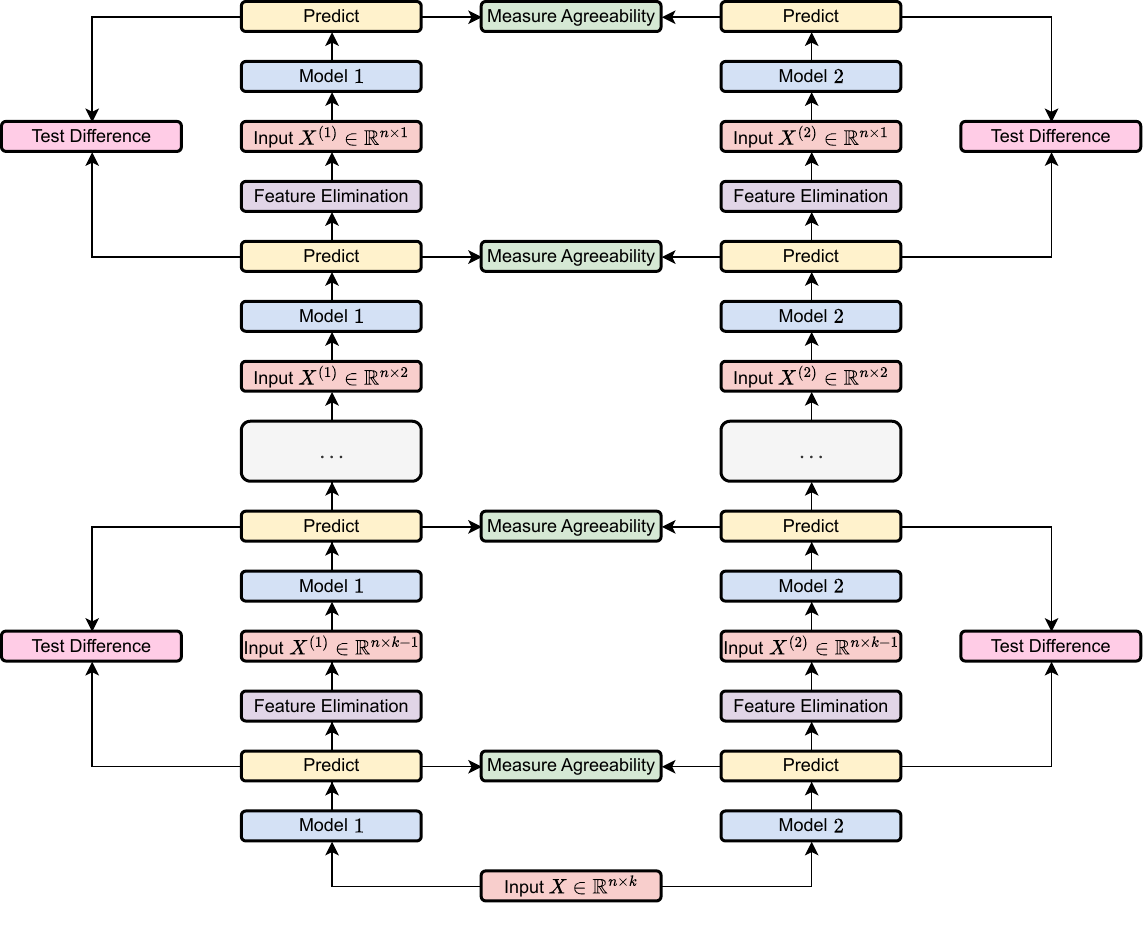}
  \caption{The entire Automated Learning for Insightful Comparison and Evaluation (ALICE) framework.}
  \label{fig:alice}
\end{figure}

The general intuition and operations given under the framework are best demonstrated in Figure~\ref{fig:alice}, which encapsulates the entire procedure of combining feature selection and inter-rater agreeability. Given some input feature set \(X \in \mathbb{R}^{n \times k}\), and given two models, here marked as Model \(1\) and Model \(2\), the framework iteratively makes predictions with each model at every stage of their respective feature elimination. At each stage of feature elimination, the predictions of each model are then sorted in accordance with best-to-worst predictive performance, and agreeability measurements are made step-by-step from between best-to-best to worst-to-worst predictions. At the same time, in both models, the features the removal of which led to most \emph{improved} or least \emph{diminished} performance are dropped from the next iteration. The process continues until the feature set is exhausted for both models. On the side, another component of the framework tests the statistical difference between some \(n\) --- or for that matter every --- best prediction sets per iteration \emph{within} models.

\subsection{Main Modules}
\label{subsec:alice_modules}

In practice, every component of the framework is wrapped up in a python class object called \texttt{BackEliminator}, which includes two distinct methods (among others) --- one for comparing two models and one for comparing results already obtained within models. A more detailed description of the algorithm will follow. In the meantime, we can proceed with describing all the modules and components of the framework at its early development stage which are:

\begin{itemize}
    \item \texttt{agreeability}
    \item \texttt{metrics}
    \item \texttt{testing}
    \item \texttt{utils}
    \item \texttt{search\_and\_compare}
\end{itemize}

\paragraph{Agreeability} This \href{https://github.com/anasashb/aliceHU/tree/main/alice/agreeability}{module} of the framework includes functions that compute coefficients of agreement between the models. There are two sets of agreeability metrics --- for classification and regression tasks. At this stage, for classification the only natively available metric  is Cohen's \(\kappa\), and for regression Pearson's \(\rho\). But in essence, this list can be expanded with ease in a locally copied version of the repository with any existing metrics within other libraries such as \texttt{scikit-learn}.

\paragraph{Metrics} The \texttt{metrics} \href{https://github.com/anasashb/aliceHU/tree/main/alice/metrics}{module} carries two sets of metrics for classification and regression. These are used to evaluate models, as well as to decide which features to eliminate. Natively implemented functions for classification are accuracy, precision, recall, and the \(F1\) score (refer to \citet{christen_review_2023} for a good review on the subject).  For regression these are mean squared error, root mean squared error, and mean absolute error. But again as with agreeability functions, this list can in future be easily expanded with implementations imported from other existing libraries. 

\paragraph{Testing} The \texttt{testing} \href{https://github.com/anasashb/aliceHU/tree/main/alice/testing}{module} for classification tasks carries natively developed code for building a McNemar's contingency table used in two wrapper functions of \texttt{statsmodels} implementation McNemar's test of homogeneity \citep{mcnemar_note_1947} with binomial and \(\chi^{2}\) distributions. For a regression task, the framework has a wrapper function of \texttt{scipy.stats} implementation of the t-test. The function, when called conducts the Levene's test \citep{brown_robust_1974} to automatically determine whether to use Student's \citep{student_probable_1908} or Welch's \citep{welch_generalization_1947} t-test.

\paragraph{Utils} The \href{https://github.com/anasashb/aliceHU/tree/main/alice/utils}{module} \texttt{utils} carries two crucial components that enable the smooth operation of the framework on the one hand with one-hot-encoded variables and on the other hand with neural networks. Firstly, the sub-module \texttt{feature\_lists} given under the module carries two relevant methods --- \texttt{dummy\_grouper} and \texttt{feature\_fixer}. The former method ensures that the user is able to provide (following the given documentation) a list of one-hot-encoded variables that were obtained from a categorical variable and thus need to be treated as \emph{one feature} during the feature elimination process. The latter method, on its part, allows users to provide a list of variables that will be fixed in the feature sets and thus be exempt from elimination at any iteration.

But a more important \href{https://github.com/anasashb/aliceHU/blob/main/alice/utils/model_training.py}{sub-module} is \texttt{model\_training} which ensures the smooth and user-friendly inclusion of neural networks in the framework. An issue of user-friendliness first arises when it comes to the training arguments for the models. At its simplest, \texttt{scikit-learn} models require essentially two arguments for fitting, the \(X\), and \(y\), which is obviously not the case for neural networks. Since the framework is tailored for use with \texttt{Keras} models, the \texttt{model\_training} sub-module includes class \texttt{KerasParams} which can be imported ahead of running the comparison algorithm. The class takes in the following arguments: training batch size (\texttt{batch\_size}) training epochs (\texttt{epochs}), validation split ratio (\texttt{validation\_split}), \texttt{callbacks} to be able to include early stopping, and \texttt{verbose} which controls the verbosity of training epoch logs. The class then outputs a parameter dictionary that can be provided to the framework at initialization and will be implicitly carried over to every round of model fitting within the algorithm. 

Another, more problematic issue associated with \texttt{Keras} neural networks as regards inclusion in the \texttt{ALICE} framework are the input shapes, which need to be pre-defined as a model is constructed and compiled. For obvious reasons, this conflicts with the inclusion in the feature elimination algorithm, where the input shape along the feature axis gradually decreases and may vary even more if several dummy variables are being treated as one feature. To preserve the simplicity of the \texttt{Sequential} API for building models in \texttt{Keras} while mitigating the requirement of predefined input shapes, the framework includes a \texttt{KerasSequential} wrapper class which allows for building a neural network with the exact syntax of the original \texttt{Sequential} API. The only key difference is the absence of an input shape argument or an input layer. The user essentially specifies the desired architecture and then the input shape gets inferred by the \texttt{KerasSequential} class iteratively during the implementation of the \texttt{BackEliminator}. Other components of the \texttt{utils} module are not user-facing and therefore we'll skip over discussing them.

\paragraph{search\_and\_compare} The \texttt{search\_end\_compare} \href{https://github.com/anasashb/aliceHU/blob/main/alice/search_and_compare/sequential.py}{module} carries the most important part of the framework, the \texttt{BackEliminator} class (refer to the Python documentation string in Appendix~\ref{sec:appendix_backeliminator}) which handles all the functionalities of the framework. The methods united under the class are as follows:

\begin{itemize}
    \item \texttt{compare\_models} --- the algorithm combining feature selection and inter-rater agreeability
    \item \texttt{compare\_n\_best} --- compares the top-\(n\) results from each models \emph{within} models based on statistical tests
    \item \texttt{dataframe\_from\_results} --- generates a \texttt{pandas DataFrame} from the results of \texttt{compare\_models} algorithm
    \item \texttt{plot\_best} --- plots best results from each iteration in the algorithm
    \item \texttt{plot\_all} --- plots all (mean) results from each iteration in the algorithm
    \item \texttt{interactive\_plot} --- generates an interactive plot of results from the algorithm
\end{itemize}

How the main algorithm of our framework, \texttt{compare\_models} (also refer to the Python documentation string in Appendix~\ref{sec:appendix_compare_models}), works can be better described in the pseudocode on Algorithm~\ref{alg:backelimination}. 

\begin{algorithm}[ht]
\caption{\texttt{compare\_models} method of the \texttt{BackEliminator} in \texttt{ALICE}}
\label{alg:backelimination}
\SetAlgoLined
\KwIn{
    \texttt{m1}, \texttt{m2} - two models for comparison;\\
    \texttt{X\_train}, \texttt{y\_train}, \texttt{X\_val}, \texttt{y\_val} - full set of predictors and targets;\\
    \texttt{keras\_params} [optional] - parameters for training a \texttt{Keras} Neural Network model. 
}

\KwOut{
    Results dictionary with performance metrics of and agreeability scores between \texttt{m1}, \texttt{m2};\\
    Feature selection logs.
}

\BlankLine
\tcp{Step 1: Fit both models with a full set of features}
fit \texttt{m1}, \texttt{m2}\;
predict on validation data with \texttt{m1}, \texttt{m2}\;
compute predictive performance scores for \texttt{m1}, \texttt{m2}\;
compute agreeability coefficients between \texttt{m1}, \texttt{m2} predictions\;

\BlankLine

\tcp{Step 2: Begin feature elimination}
copy the original feature set and assign to \texttt{m1}, \texttt{m2}\;

\BlankLine
\While{both models' feature sets are not empty}{
    \For{each model}{
        \For{each feature in the model's current feature set}{
            remove feature\;
            fit model on remaining set\;
            predict on validation data\;
            compute predictive performance scores\;
        }
    }    
    sort \texttt{m1}, \texttt{m2} scores from best to worst\;
    \For{score in sorted \texttt{m1} scores, sorted \texttt{m2} scores}{
        compute agreeability between \texttt{m1} and \texttt{m2} predictions based on score order\;
    }
    report agreeability between \texttt{m1}, \texttt{m2} best predictions\;
    report mean agreeability between \texttt{m1}, \texttt{m2} and standard deviation\;
    pick worst performing feature for \texttt{m1}, \texttt{m2} based on best predictions\;
    update \texttt{m1} feature set, \texttt{m2} feature set for next iteration\;   
\BlankLine
\tcp{Iteration end}
print iteration summary\;
\tcp{Move to next iteration}
\BlankLine
}
\Return{results dictionary}

\end{algorithm}

The algorithm not only allows the selection of an optimal set of features using backward elimination, but it also helps provide a significant insight into how two compared models fare when the input feature sets are manipulated. A more important aspect of the framework, however, is the generation of agreeability metrics between the two best predictions from both included models at each iteration of the elimination. In a specific, data-driven predictive modeling scenario, where a user is evaluating a white-box and a black-box model, such functionality of the framework allows for getting a general idea of the trade-off between optimal predictive performance and interpretability by being able to inspect which subset of features leads to the best prediction by, for example, a neural network \emph{in terms of error metrics} and the best prediction \emph{in terms of agreeability} with a white-box model.

The second key component of the framework is the \texttt{compare\_n\_best} method (refer to the Python documentation string in Appendix~\ref{sec:appendix_compare_n_best}), already mentioned above. This function requires the user to have already ran the \texttt{compare\_models} algorithm and obtained the results. Based on the results, as depicted in Figure~\ref{fig:alice}, each model's best predictions at each elimination iteration are compared \emph{within} models. In other words, given model \(A\), model \(B\), and their respective predictions, the function will conduct tests to determine whether 1\textsuperscript{st} and 2\textsuperscript{nd} best predictions from model \(A\) are statistically significantly different from each other, and the same for the model \(B\). The number of \(n\)-best predictions to compare can be picked by the user. This serves as a further sneak peek of insight into the models' doings by helping to determine how variability-prone each model included in an experiment is to the given data.

As regards the rest of the methods included in the \texttt{BackEliminator}, we believe the small descriptions given in the list above can suffice. A better demonstration of each functionality can be observed with the results given in Section~\ref{sec:results} as well as in the experiment \href{https://github.com/anasashb/aliceHU/blob/main/customer_churn_test.ipynb}{demo notebook} in the \texttt{ALICE} GitHub repository.

\section{Experiment Setting}
\subsection{Data}
\label{subsec:exp_data}
The data used in the experiments is the Telco Customer Churn dataset by IBM\footnote{Available on \href{https://www.kaggle.com/datasets/blastchar/telco-customer-churn}{Kaggle} and in the \texttt{ALICE} \href{https://github.com/anasashb/aliceHU/blob/main/Telco_customer_churn.xlsx}{GitHub repository}.} which has been commonly used in feature selection experiments \citep{yulianti_sequential_2020, wu_integrated_2021, sana_novel_2022}. The target is whether the customer left the telecommunications company in the given quarter. There are \(7,043\) observations, of which we keep \(7,032\) after cleaning. The dataset includes \(32\) raw variables, of which we keep \(22\) (refer to Table~\ref{tab:feature_set} in Appendix~\ref{sec:appendix_feature_set} for the list and variable descriptions). With one-hot-encoding of categorical features, we end up with \(32\) predictors in total. One-hot-encoding was picked for categorical variables over more complex or sophisticated data transformation methods due to its relative simplicity, as the aim of the experiments was not to necessarily squeeze out the best possible performance out of our classifiers but to compare their inner workings. Our train-test split is of \(0.8\)-\(0.2\) ratio. Due to a severe imbalance in the customer churn variable, we use the Synthetic Minority Over-Sampling Technique (SMOTE; \citep{chawla_smote_2002}) on the training set. The entire \href{https://github.com/anasashb/aliceHU/blob/main/customer_churn_dataprocessing.ipynb}{data cleaning and processing notebook}, as well as the \href{https://github.com/anasashb/aliceHU/tree/main/clean_data/class_telco}{cleaned datasets}, are disclosed in the \texttt{ALICE} repository for transparency purposes.

\subsection{Models}
\label{subsec:exp_models}
Three models included in the experiments are a Logistic Regression, a Random Forest Classifier, and a Feed-Forward Neural Network / Multi-Layer Perceptron\footnote{Will be abbreviated as MLP for the rest of the paper}. The Logit is implemented with \texttt{scikit-learn} using the \texttt{liblinear} solver and an \(L2\) (Ridge) penalty. The RFC is also a standard implementation using \texttt{scikit-learn} as an ensemble of 100 estimators, with no maximum depth specified and a minimum number of \(2\) samples required to split a node. The split criterion is Gini impurity. 

\begin{figure}[h] 
  \centering
  \includegraphics[width=\textwidth]{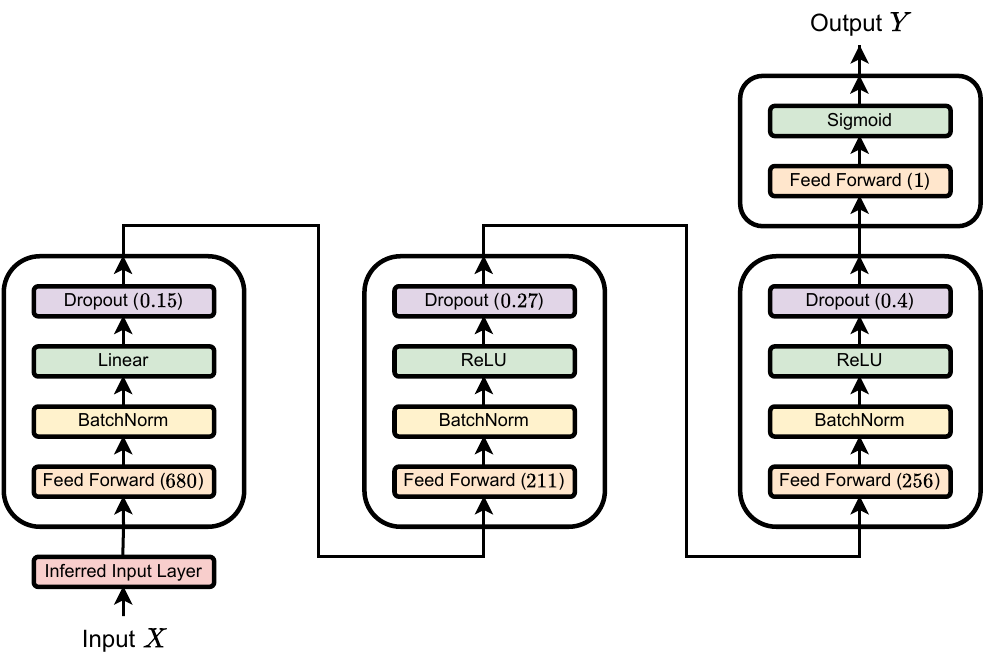}
  \caption{Deep Feed-Forward Network obtained via Differential Evolution-based hyperparameter optimization.}
  \label{fig:simple_mlp}
\end{figure}

The MLP for our experiments is a bit more complex in architecture (see Figure~\ref{fig:simple_mlp}). Implemented with \texttt{TensorFlow.keras}, it has three main blocks with fully connected layers, Batch Normalization \citep{ioffe_batch_2015} and linear and Rectified Linear Unit (ReLU; \citep{fukushima_neocognitron_1980}) activation functions, and a fourth block for output projection with a sigmoid head. The number of layers and units in the three blocks were tuned using an experimental Python implementation of the Differential Evolution algorithm \citep{storn_differential_1997} for hyper-parameter optimization \citep{capone_vincaptheotf_2022}.

The training parameters for the neural network classifier in all experiments were as follows: the batch size of \(32\), \(100\) training epochs, further training-validation split of \(20\%\), and early stopping of training after no observed improvement in validation performance after \(3\) epochs. Adaptive Moment Estimation (Adam; \citep{kingma_adam_2015}) was used as the optimizer and binary cross entropy was used as the loss function (refer to \citet{chen_chapter_2024} for an overview of the function and its \texttt{TensorFlow} implementation). 

As an evaluation metric of the models, the \(F1\) score was picked. The same metric was also used to score the feature elimination process. The classification threshold was kept at a standard \(\tau = 0.5\) across all models in all experiments. 

\section{Results}
\label{sec:results}

Given the dataset described in Section~\ref{subsec:exp_data} and the models covered in Section~\ref{subsec:exp_models}, we ran three experiments in total where we provided for comparison to the \texttt{BackEliminator}: a) MLP and a Logistic Regression; b) MLP and a Random Forest Classifier; c) Logistic Regression and Random Forest Classifier. Given the feature count (treating dummies obtained from a single categorical variable as one) of \(22\), there were \(22\) iterations of the \texttt{BackEliminator} in each of the experiments. Cross-comparing the models in different experiments also allowed us to gauge how robust the non-parametric models were across trials. 

\begin{centering}    
\begin{longtable}{@{}c|ccccccccc@{}}
\caption{Best and mean agreeability scores from each trial.}
\label{tab:main_results} \\
\toprule
\multirow{2}{*}{\rotatebox[origin=c]{90}{Iter.}} & \multicolumn{3}{c|}{\textbf{MLP vs. Logit}} & \multicolumn{3}{c|}{\textbf{MLP vs. RFC}} & \multicolumn{3}{c}{\textbf{Logit vs. RFC}} \\
 

&
\multicolumn{1}{c}{Best} & \multicolumn{1}{c}{Mean} & \multicolumn{1}{c|}{St. Dev.} &  
\multicolumn{1}{c}{Best} & \multicolumn{1}{c}{Mean} & \multicolumn{1}{c|}{St. Dev.} & 
\multicolumn{1}{c}{Best} & \multicolumn{1}{c}{Mean} & \multicolumn{1}{c}{St. Dev.} \\

\midrule

\endfirsthead

\multicolumn{10}{c}{Table \thetable\ continued from previous page} \\

\toprule

\multirow{2}{*}{\rotatebox[origin=c]{90}{Iter.}} & \multicolumn{3}{c|}{\textbf{MLP vs. Logit}} & \multicolumn{3}{c|}{\textbf{MLP vs. RFC}} & \multicolumn{3}{c}{\textbf{Logit vs. RFC}} \\


&
\multicolumn{1}{c}{Best} & \multicolumn{1}{c}{Mean} & \multicolumn{1}{c|}{St. Dev.} &  
\multicolumn{1}{c}{Best} & \multicolumn{1}{c}{Mean} & \multicolumn{1}{c|}{St. Dev.} & 
\multicolumn{1}{c}{Best} & \multicolumn{1}{c}{Mean} & \multicolumn{1}{c}{St. Dev.} \\

\midrule

\endhead

\multicolumn{1}{c|}{1}   &  \textbf{0.846} &	-              &    \multicolumn{1}{c|}{-}      & 0.796	& -     & \multicolumn{1}{c|}{-}     & 0.791 & -     & -     \\
\multicolumn{1}{c|}{2}   &  \textbf{0.858} &	\textbf{0.837} &	\multicolumn{1}{c|}{0.026}	& 0.772 & 0.774 & \multicolumn{1}{c|}{0.021} & 0.795 & 0.772 & 0.024 \\
\multicolumn{1}{c|}{3}   &  \textbf{0.861} &	\textbf{0.843} &    \multicolumn{1}{c|}{0.027}	& 0.780	& 0.766 & \multicolumn{1}{c|}{0.021} & 0.787 & 0.752 & 0.023 \\
\multicolumn{1}{c|}{4}   &  \textbf{0.822} &	\textbf{0.819} &	\multicolumn{1}{c|}{0.023}  & 0.782 & 0.755	& \multicolumn{1}{c|}{0.026} & 0.787 & 0.749 & 0.023 \\
\multicolumn{1}{c|}{5}   &  \textbf{0.813} &	\textbf{0.813} &	\multicolumn{1}{c|}{0.037}	& 0.782	& 0.767	& \multicolumn{1}{c|}{0.028} & 0.767 & 0.751 & 0.034 \\
\multicolumn{1}{c|}{6}   &  \textbf{0.857} &	\textbf{0.807} &	\multicolumn{1}{c|}{0.020}	& 0.754	& 0.750	& \multicolumn{1}{c|}{0.024} & 0.752 & 0.732 & 0.037 \\
\multicolumn{1}{c|}{7}   &  \textbf{0.879} &	\textbf{0.825} &	\multicolumn{1}{c|}{0.036}  & 0.765	& 0.750	& \multicolumn{1}{c|}{0.033} & 0.750 & 0.725 & 0.041 \\
\multicolumn{1}{c|}{8}   &	\textbf{0.857} &	\textbf{0.836} &	\multicolumn{1}{c|}{0.037}	& 0.765	& 0.755	& \multicolumn{1}{c|}{0.024} & 0.752 & 0.731 & 0.040 \\
\multicolumn{1}{c|}{9}   &	\textbf{0.850} &	\textbf{0.827} &	\multicolumn{1}{c|}{0.039}	& 0.750	& 0.738	& \multicolumn{1}{c|}{0.028} & 0.737 & 0.723 & 0.040 \\
\multicolumn{1}{c|}{10}  &	\textbf{0.859} &	\textbf{0.814} &	\multicolumn{1}{c|}{0.032}	& 0.777	& 0.738	& \multicolumn{1}{c|}{0.027} & 0.724 & 0.710 & 0.044 \\
\multicolumn{1}{c|}{11}  &	\textbf{0.810} &	\textbf{0.822} &	\multicolumn{1}{c|}{0.033}	& 0.782	& 0.738	& \multicolumn{1}{c|}{0.035} & 0.716 & 0.693 & 0.045 \\
\multicolumn{1}{c|}{12}  &	\textbf{0.811} &	\textbf{0.787} &	\multicolumn{1}{c|}{0.038}	& 0.763	& 0.734	& \multicolumn{1}{c|}{0.035} & 0.736 & 0.684 & 0.038 \\
\multicolumn{1}{c|}{13}  &	\textbf{0.806} &    \textbf{0.772} &	\multicolumn{1}{c|}{0.050}	& 0.761	& 0.725	& \multicolumn{1}{c|}{0.038} & 0.746 & 0.701 & 0.044 \\
\multicolumn{1}{c|}{14}  &	\textbf{0.810} &	\textbf{0.784} &	\multicolumn{1}{c|}{0.051}	& 0.736	& 0.713	& \multicolumn{1}{c|}{0.035} & 0.741 & 0.712 & 0.034 \\
\multicolumn{1}{c|}{15}  &	0.761 &	0.780 &	\multicolumn{1}{c|}{0.044}	& 0.722	& 0.710	& \multicolumn{1}{c|}{0.035} & 0.770 & 0.694 & 0.056 \\
\multicolumn{1}{c|}{16}  &	0.786 &	0.717 &	\multicolumn{1}{c|}{0.062}	& 0.776	& 0.673	& \multicolumn{1}{c|}{0.061} & 0.769 & 0.706 & 0.049 \\
\multicolumn{1}{c|}{17}  &	0.676 &	0.715 &	\multicolumn{1}{c|}{0.047}	& 0.768	& 0.730	& \multicolumn{1}{c|}{0.055} & 0.700 & 0.714 & 0.040 \\
\multicolumn{1}{c|}{18}  &	0.684 & 0.608 &	\multicolumn{1}{c|}{0.053}	& 0.711	& 0.694	& \multicolumn{1}{c|}{0.046} & 0.704 & 0.599 & 0.067 \\
\multicolumn{1}{c|}{19}  &	0.577 &	0.597 &	\multicolumn{1}{c|}{0.053}	& 0.665	& 0.565	& \multicolumn{1}{c|}{0.070} & 0.610 & 0.576 & 0.060 \\
\multicolumn{1}{c|}{20}  &	0.617 &	0.474 &	\multicolumn{1}{c|}{0.113}	& 0.440	& 0.437	& \multicolumn{1}{c|}{0.098} & 0.511 & 0.449 & 0.092 \\
\multicolumn{1}{c|}{21}  &	0.458 &	0.357 &	\multicolumn{1}{c|}{0.138}	& 0.306	& 0.309	& \multicolumn{1}{c|}{0.016} & 0.398 & 0.246 & 0.153 \\
\multicolumn{1}{c|}{22}  &	0.453 & 0.143 & \multicolumn{1}{c|}{0.309}	& 0.470	& 0.353	& \multicolumn{1}{c|}{0.118} &0.540	 & 0.214 & 0.327 \\

\bottomrule

\end{longtable}
\end{centering}

Results shown in Figure~\ref{tab:main_results} seem to suggest that the MLP classifier and the Logistic Regression, compared to the other combinations maintain almost perfect agreement given a sufficient amount of features. On the specific task, between the two models, the agreeability only takes a fall when less than \(9\) features are included in each model. This finding is supported by the reported mean agreeability, which is the average of measurements taken during a single iteration of the feature elimination algorithm when features are being deselected individually to then pick which one to drop after the iteration. 

It can also be observed that the MLP and the RFC or the Logit and the RFC, in their respective experiments, do not surpass the level of substantial agreement under any combinations of the input features. Besides, the results show that when given less than five features, the best agreeability scores in all experiments are either very close to \(0.6\) or fall below it, indicating merely a moderate agreement between the models. This is even more pronounced in mean scores and their corresponding widening standard deviations.

While it is hard to make an argument that is not complete speculation, still the finding that the neural network and the Logistic Regression have high agreement may not come as a surprise given the fact that the MLP \emph{does} have a sigmoid head and \emph{is} optimized with a gradient-based method similarly to the Logit despite the significant difference in overall trainable parameters in both of the models. 

\begin{figure}[H] 
  \centering
  \includegraphics[width=\textwidth]{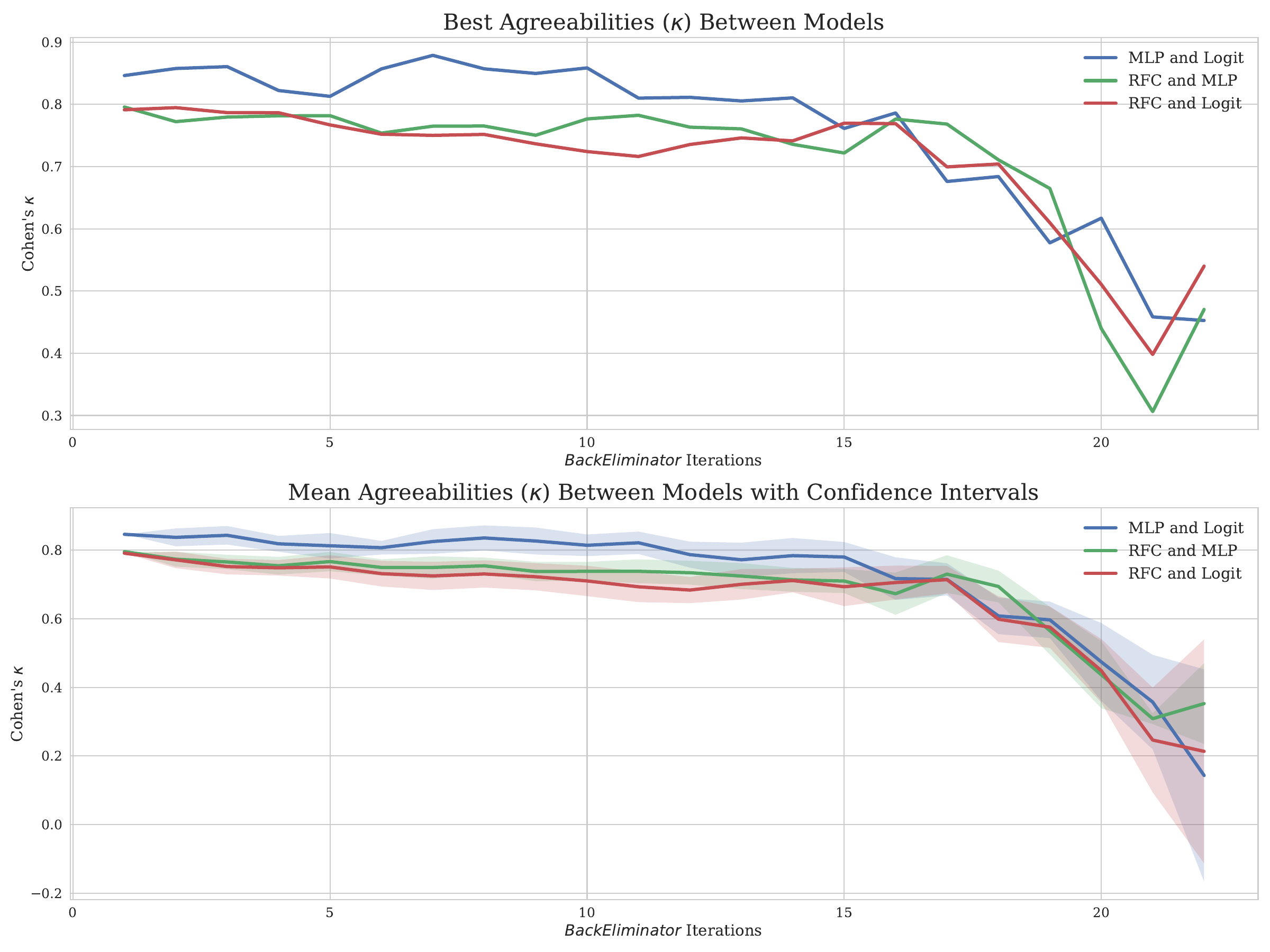}
  \caption{Best and mean agreeabilities between models across the three experiments.}
  \label{fig:model_to_model}
\end{figure}

Best and mean agreeability scores from the three experiments can also be observed in Figure~\ref{fig:model_to_model}, which also confirms that the higher agreement between MLP and Logit is not at random given the confidence intervals depicted using the standard deviations from the mean agreeability coefficients. While the predictive performance measures are not by themselves as informative for the purpose of our experiment, a more detailed look at the results from each trial where the agreeability is presented alongside \(F1\) scores of each model can be inspected in the Appendix~\ref{sec:appendix_mlp_logit} for MLP and Logit, Appendix~\ref{sec:appendix_mlp_rfc} for MLP and RFC, and Appendix~\ref{sec:appendix_logit_rfc} for Logit and RFC. 

\begin{centering}    
\begin{longtable}{@{}c|cccccc@{}}
\caption{McNemar's test statistics between the \(5\) best predictions from each model in all experiments.}
\label{tab:n_best_results} \\
\toprule
\multirow{2}{*}{\(n\)-best} & \multicolumn{2}{c|}{\textbf{MLP vs. Logit}} & \multicolumn{2}{c|}{\textbf{MLP vs. RFC}} & \multicolumn{2}{c}{\textbf{Logit vs. RFC}} \\
 

&
\multicolumn{1}{c}{MLP}   & \multicolumn{1}{c|}{Logit} &  
\multicolumn{1}{c}{MLP}   & \multicolumn{1}{c|}{RFC}   &  
\multicolumn{1}{c}{Logit} & \multicolumn{1}{c}{RFC}    \\

\midrule

\endfirsthead

\multicolumn{7}{c}{Table \thetable\ continued from previous page} \\

\toprule

\multirow{2}{*}{\(n\)-best} & \multicolumn{2}{c|}{\textbf{MLP vs. Logit}} & \multicolumn{2}{c|}{\textbf{MLP vs. RFC}} & \multicolumn{2}{c}{\textbf{Logit vs. RFC}} \\


&
\multicolumn{1}{c}{MLP}   & \multicolumn{1}{c|}{Logit} &  
\multicolumn{1}{c}{MLP}   & \multicolumn{1}{c|}{RFC}   &  
\multicolumn{1}{c}{Logit} & \multicolumn{1}{c}{RFC}    \\

\endhead

\multicolumn{1}{c|}{1\textsuperscript{st} vs. 2\textsuperscript{nd}}  & 115.7 & \multicolumn{1}{c|}{283.1} & 182.1 & \multicolumn{1}{c|}{152.7} & 283.1 & \multicolumn{1}{c}{186.2} \\
\multicolumn{1}{c|}{2\textsuperscript{nd} vs. 3\textsuperscript{rd}}  & 171.0 & \multicolumn{1}{c|}{277.3} & 109.7 & \multicolumn{1}{c|}{140.5} & 277.2 & \multicolumn{1}{c}{172.7} \\
\multicolumn{1}{c|}{3\textsuperscript{rd} vs. 4\textsuperscript{th}}  & 124.2 & \multicolumn{1}{c|}{204.0} & 188.8 & \multicolumn{1}{c|}{151.9} & 204.0 & \multicolumn{1}{c}{187.2} \\
\multicolumn{1}{c|}{4\textsuperscript{th} vs. 5\textsuperscript{th}}  & 234.5 & \multicolumn{1}{c|}{204.7} & 164.8 & \multicolumn{1}{c|}{170.6} & 204.7 & \multicolumn{1}{c}{167.7} \\

\bottomrule

\end{longtable}
\end{centering}

The success of the second, \texttt{compare\_n\_best}, method is not as evident at first glance. Table~\ref{tab:n_best_results} shows the McNemar's \(\chi^2\) test results for \(5\)-best models in each experiment. The p-values are not reported as all of them were \(0.000000\) at the precision of the sixth decimal. While this shows that for all models the predictions were statistically significantly different from each other, there could be a take-away from the test statistic scores themselves, which are reported in the table. Firstly, it is shown clearly how robust the Logistic Regression is as its test statistics are extremely close to each other in the MLP vs. Logit and Logit vs. RFC experiments. Besides it can be seen that the statistic does not vary as much as that of MLP or the RFC per experiment, and is more stable for example from the best Logit model to the second best Logit model than when it comes to the best neural network predictions to the second best neural network predictions.

Besides, the robustness of the Logistic Regression is clearly displayed when comparing its feature elimination steps and achieved \(F1\) scores in different experiments: MLP vs. Logit (see Table~\ref{tab:mlp_logit_results} in Appendix~\ref{sec:appendix_mlp_logit}) and RFC vs. Logit (see Table~\ref{tab:rfc_logit_results} in Appendix~\ref{sec:appendix_logit_rfc}). Unlike the Random Forest Classifier and the Multi-Layer Perceptron, the features dropped by the Logistic Regression are the same iteration-for-iteration in the two experiments that included it, and the best performance is also achieved with the same sets of features in both. This is clearly not the case for the two experiments that included the neural network: MLP vs. Logit, and MLP vs. RFC (see Table~\ref{tab:rfc_mlp_results} in Appendix~\ref{sec:appendix_mlp_rfc}), where the order of features eliminated of the MLP varied, and best performance in terms of \(F1\) scores was achieved on different subsets of features. In the three table results it can also be verified that the feature subsets that led to these models' best agreeabilities were not the same that led to best performance. This again confirms what was already discussed in terms of the trade-off between performance and agreeability, and having access to such results would allow users to make more informed decisions which model to choose for deployment following evaluations for their own specific modeling task. 

\section{Conclusion and Outlook}
\label{sec:conclusion}
Over the scope of the paper, we have demonstrated a novel framework that combines the basics of feature selection and inter-rater agreeability for obtaining a deeper look into the inner workings of machine learning models. Our analysis provides valuable insights into the realm of ML algorithms, although their inherent complexity often renders them as black boxes. While Logistic Regression stands out for its comprehensibility, comparing other algorithms to it offers a means to glean insights. This comparative approach, achieved through rigorous statistical evaluations using diverse input variables representing varying degrees of information, extends to any comparison between two ML algorithms. Furthermore, our analysis underscores the importance of assessing the robustness of these algorithms, particularly in practical contexts where data volumes are substantial. Given the necessity of selecting input variables when dealing with big data, understanding the
robustness of algorithms becomes crucial. Such insights enable practitioners to make informed decisions regarding model specifications, thereby enhancing the likelihood of identifying optimal solutions.

 While the results from initial experiments seem to show promise, there is also much room for extending the framework and the scope of experiments. The framework in its current development stage fully supports \emph{regression tasks}, which we have not experimented with yet. There is also room for expanding the \texttt{search\_and\_compare} module to include other conventional algorithms such as Forward Selection and Plus-\(l\)-Take-Away-\(r\) (see Section~\ref{subsec:feature_selection}). Other coefficients for agreeability could also be implemented, including variations of Cohen's \(\kappa\) itself. Another possibility for the future would be to implement a Recursive Feature Elimination \citep{guyon_gene_2002} algorithm where the features from a neural network are eliminated based on the coefficients of a logit or a linear model and see the agreeability there. Another interesting avenue to test the framework would be using benchmark datasets for feature selection from the field of medicine and genetics, or expanding existing benchmarks with synthetic, non-sensical features to see how robust the given models are for eliminating (or not selecting) them, and to what extent they agree on this decision. From the standpoint of supported libraries, there is also room for expansion --- in theory, the framework should be easily adaptable to support ML models and their corresponding Python libraries that are outside the \texttt{scikit-learn} ecosystem but still use the \texttt{.fit()}; \texttt{.predict()} syntax such as eXtreme Gradient Boosting (\texttt{XGBoost}; \citep{chen_xgboost_2016}) and \texttt{CatBoost} \citep{dorogush_catboost_2018}.

\bibliographystyle{apalike}  
\bibliography{references}  

\newpage
\section*{Appendix}
\appendix
\section{\texttt{BackEliminator} Docstring}
\label{sec:appendix_backeliminator}

\begin{verbatim}
class BackEliminator:
    """
    The class is built for conducting backwards feature elimination in
    combination with model agreeability.

    Args:
        X (pd.DataFrame): A pandas dataframe containing predictors.
        y (pd.DataFrame): A pandas dataframe containing target.
        validation_data (tuple): A tuple of validation data
                                 (X_val, y_val).
        task_type (str): String for task type. Available options -
                         "classification" or "regression".
        criterion (str): String for intra-model evaluation criterion.
                         Available options: ("mse", "rmse", "mae",
                                            "accuracy", "precision",
                                            "recall", "f1")
        agreeability (str): String for inter-model comparison.
                            Available options: "pearson", "cohen_kappa"
        dummy_list (list): List of lists containing column names (str)
                           of dummy features generated from a
                           categorical variable. (Optional).
        features_to_fix (list): List containing column names (str) of
                                features that will be excluded from
                                feature elimination and thus always
                                included in modeling. (Optional)

    >>> Regression Example:
        seeker = BackEliminator(
            X=X_train,
            y=y_train,
            validation_data=(X_val, y_val),
            task_type="regression",
            criterion="rmse",
            agreeability="pearson",
            dummy_list=[
                ["dummy_1_from_variable_1", "dummy_2_from_variable_1"],
                [
                    "dummy_1_from_variable_2",
                    "dummy_2_from_variable_2",
                    "dummy_3_from_variable_2",
                ],
            ],
            features_to_fix=[
                "variable_3",
                "variable_4",
            ],
        )

    >>> Classification Example:
        seeker = Backeliminator(
            X=X_train,
            y=y_train,
            validation_data=(X_val, y_val),
            task_type="classification",
            criterion="f1",
            agreeability="cohen_kappa",
            dummy_list=[
                ["dummy_1_from_variable_1", "dummy_2_from_variable_1"],
                [
                    "dummy_1_from_variable_2",
                    "dummy_2_from_variable_2",
                    "dummy_3_from_variable_2",
                ],
            ],
            features_to_fix=[
                "variable_3",
                "variable_4",
            ]
        )
    """    
\end{verbatim}

\section{\texttt{compare\_models} Docstring}
\label{sec:appendix_compare_models}
\begin{verbatim}
def compare_models(self, m1, m2, keras_params=None):
    """
    Fits and evaluates two different models with various subsets of
    features. Measures inter-rater agreeability between models'
    predictions on the validation/test set.

    Args:
        m1: Sklearn or Keras model.
        m2: Sklearn or Keras model.
        keras_params (optional): KerasParams object carrying
                                 pre-defined configuration for
                                 training a Keras model -
                                 batch_size, epochs,
                                 validation_split, callbacks,
                                 verbose arguments do be called in
                                 training.

    >>> Note on including Keras models:
        Keras models must be provided as a KerasSequential class
        object given in this library to ensure proper compiling,
        fitting and inference during the iteration's of the
        algorithm.

    >>> Example use:
        # Define KerasSequential model
        mlp = KerasSequential()  # Initialize as KerasSequential
        mlp.add(  # 128 units, linear activation
            tf.keras.layers.Dense, units=128, activation="linear"
        )
        mlp.add(  # 64 units, linear activation
            tf.keras.layers.Dense, units=64, activation="linear"
        )
        mlp.add(  # Sigmoid output layer
            tf.keras.layers.Dense, units=1, activation="sigmoid"
        )
        mlp.compile(
            optimizer="adam",  # default lr: 0.001 for adam
            loss=tf.keras.losses.BinaryCrossentropy(),
            metrics=["accuracy"],  # Track accuracy
        )

        # Early stopping callback
        EARLY_STOPPING = [
            tf.keras.callbacks.EarlyStopping(
                monitor="val_loss",
                patience=5,
                restore_best_weights=True,
            )
        ]  
        # Define keras_params for training
        keras_params = KerasParams(
            batch_size=32,  # batch size for mini-batch training
            epochs=100,  # fit a model for 100 epochs
            validation_split=0.2,  # further 0.8-0.2 split
            callbacks=EARLY_STOPPING,  # include early stopping
            verbose=0,  # no per-epoch logs
        )

        # Define Random Forest Classifier
        rfc = RandomForestClassifier(n_estimators=100)

        # Run the algorithm via an initialized BackEliminator class
        # (refer to BackEliminator documentation)
        results = seeker.compare_models(
            m1=rfc,  # Model 1: Random Forest Classifier
            m2=mlp,  # Model 2: MultiLayer Perceptron
            keras_params=keras_params,  # Training Parameters for MLP
        )
    """
\end{verbatim}

\section{\texttt{compare\_n\_best} Docstring}
\label{sec:appendix_compare_n_best}
\begin{verbatim}
def compare_n_best(self, n=None, test=None):
    """
    Method for pair-wise comparison of n amount of best predictions
    obtained by the models. The pairwise tests are conducted within
    the predictions of each models and will test if predictions
    obtained are statistically significantly different from each
    other.

    Args:
        n (int): How many best results to compare.
        test (str): Statistical test to use. Options:
                    "mcnemar_binomial" and "mcnemar_chisquare" for
                    binary classification. "t_test" for regression.

    >>> Example: Setting n=3 will test:
                 - M1: Best predictions against second best predictions;
                       Second best predictions and third best predictions.
                 - M2: Best predictions against second best predictions;
                       Second best predictions and third best predictions.
    """
\end{verbatim}

\section{Initial Variables Included from the Telco Customer Churn}
\label{sec:appendix_feature_set}
\begin{centering}    
\begin{longtable}{@{}ccl@{}}
\caption{Full set of variables included in experiments following data cleaning.}
\label{tab:feature_set} \\
\toprule
\textbf{Feature} & \textbf{Type} & \textbf{Description} \\
\midrule
\texttt{Latitude} & Float & Latitude of customer’s residence \\ 
\texttt{Longitude} & Float & Longitude of customer’s residence \\
\texttt{TenureMonths} & Float & Total time customer has been with company \\ 
\texttt{MonthlyCharges} & Float & Customer’s total monthly charge \\
\texttt{TotalCharges} & Float & Customer’s total charges \\
\texttt{CLTV} & Float & Customer lifetime value determined by company \\
\texttt{Gender} & Binary & Customer’s gender \\
\texttt{SeniorCitizen} & Binary & Whether customer is older than 65 \\
\texttt{Partner} & Binary & Whether customer has a partner \\
\texttt{Dependents} & Binary & Whether customer lives with dependents \\
\texttt{PhoneService} & Binary & Whether customer subscribes to company home
phone service \\ 
\texttt{PaperlessBilling} & Binary & Whether customer chose paperless billing \\
\texttt{MultipleLines} & \(3\)-class & Whether customer subscribes to multiple telephone
lines \\
\texttt{InternetService} & \(3\)-class & Whether customer subscribes to internet service \\
\texttt{OnlineSecurity} & \(3\)-class & Whether customer subscribes to company’s online
security service \\
\texttt{OnlineBackup} & \(3\)-class & Whether customer subscribes to company’s online
backup service \\
\texttt{DeviceProtection} & \(3\)-class & Whether customer has device protection \\ 
\texttt{TechSupport} & \(3\)-class & Whether customer subscribes to additional customer
support service \\
\texttt{StreamingTV} & \(3\)-class & Whether customer subscribes to company’s online TV
streaming service \\
\texttt{StreamingMovies} & \(3\)-class & Whether customer subscribes to company’s movie streaming service \\
\texttt{Contract} & \(3\)-class & Customer’s contract type \\
\texttt{PaymentMethod} & \(4\)-class & How the customer pays their bill \\

\midrule
\texttt{ChurnValue} & Binary & Whether customer churned \\
\bottomrule
\end{longtable}
\end{centering}

\section{Experiment Results: MLP vs. Logit}
\label{sec:appendix_mlp_logit}
\begin{figure}[H]
  \centering
  \includegraphics[width=\textwidth]{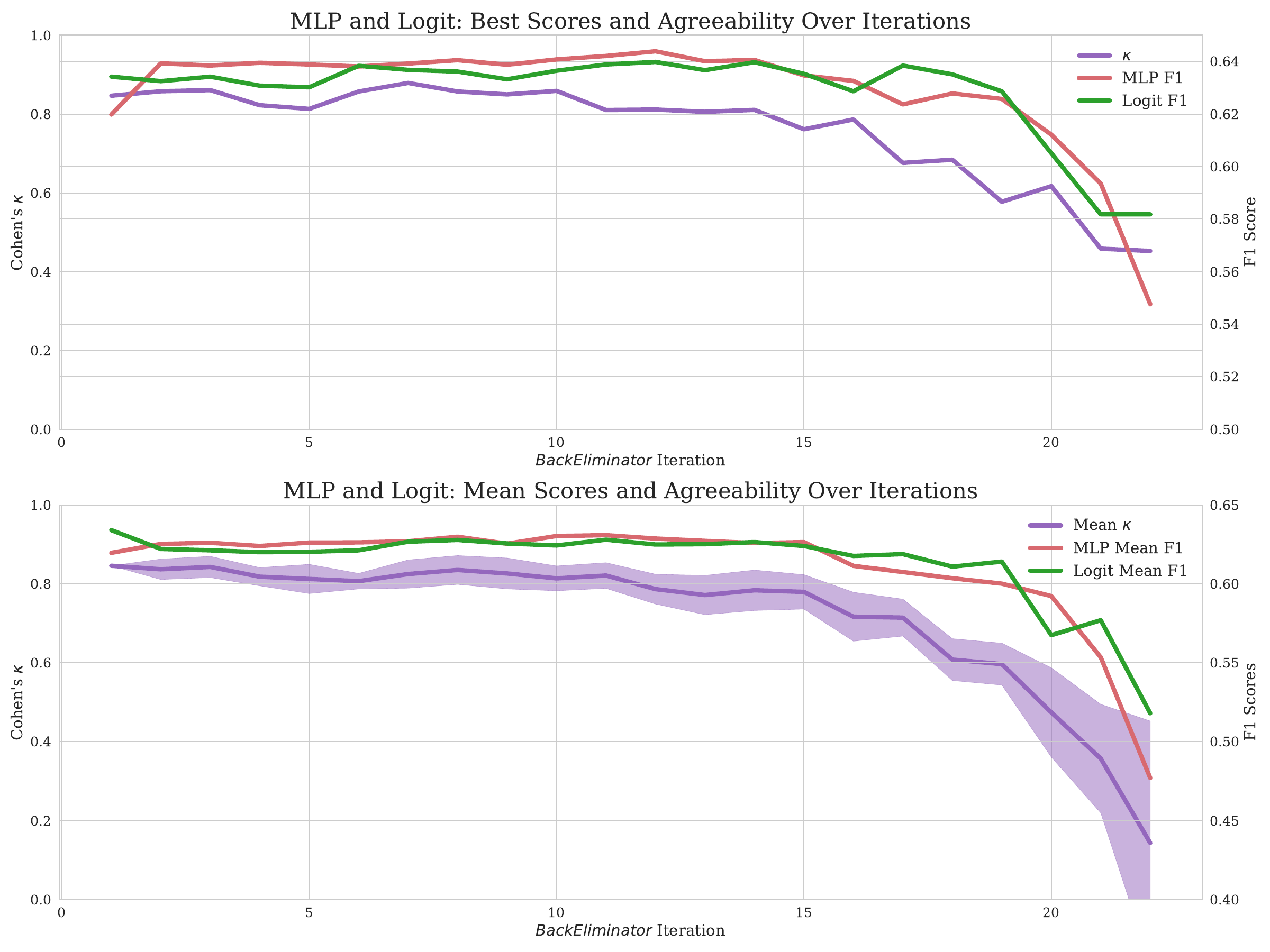}
  \caption{Best and mean \(F1\) (right-hand-side \(y\)-axis), \(\kappa\) (left-hand-side \(y\)-axis) scores from the MLP vs. Logit trial.}
  \label{fig:mlp_logit}
\end{figure}

\begin{centering}    
\begin{longtable}{@{}c|p{6cm}|c|p{6cm}|c@{}}
\caption{Included features and performance of each Model in MLP vs. Logit Trial. Bold typeface means best feature sets for \(F1\), dark green for best \(\kappa\).}
\label{tab:mlp_logit_results} \\
\toprule
\multirow{2}{*}{\rotatebox[origin=c]{90}{Iter.}} & \multicolumn{2}{c|}{\textbf{MLP}} & \multicolumn{2}{c}{\textbf{Logit}}  \\

\cline{2-5}

&
\multicolumn{1}{c|}{Features} & \multicolumn{1}{c|}{\(F1\)} & \multicolumn{1}{c|}{Features} & \multicolumn{1}{c}{\(F1\)} \\

\endfirsthead

\multicolumn{5}{c}{Table \thetable\ continued from previous page} \\

\toprule
\multirow{2}{*}{\rotatebox[origin=c]{90}{Iter.}} & \multicolumn{2}{c|}{\textbf{MLP}} & \multicolumn{2}{c}{\textbf{Logit}}  \\

\cline{2-5}

&
\multicolumn{1}{c|}{Features} & \multicolumn{1}{c|}{\(F1\)} & \multicolumn{1}{c|}{Features} & \multicolumn{1}{c}{\(F1\)} \\
\midrule

\endhead
\bottomrule
1

&

\parbox[t]{5cm}{[Latitude, Longitude, TenureMonths, MonthlyCharges, 
TotalCharges, CLTV, Gender, SeniorCitizen, 
Partner, Dependents, PhoneService, PaperlessBilling, 
MultipleLines, InternetService, OnlineSecurity, OnlineBackup, 
DeviceProtection, TechSupport, StreamingTV, StreamingMovies, 
Contract, PaymentMethod]}

&

0.620

&

\parbox[t]{5cm}{[Latitude, Longitude, TenureMonths, MonthlyCharges, TotalCharges, CLTV, Gender, SeniorCitizen, Partner, Dependents, PhoneService, PaperlessBilling, MultipleLines, InternetService, OnlineSecurity, OnlineBackup, DeviceProtection, TechSupport, StreamingTV, StreamingMovies, Contract, PaymentMethod]}

& 0.634 \\
\midrule
2	
&
\parbox[t]{5cm}{[Latitude, Longitude, TenureMonths, MonthlyCharges, TotalCharges, CLTV, Gender, SeniorCitizen, Partner, Dependents, PhoneService, PaperlessBilling, MultipleLines, OnlineSecurity, OnlineBackup, DeviceProtection, TechSupport, StreamingTV, StreamingMovies, Contract, PaymentMethod]}	
&
0.639	
&
\parbox[t]{5cm}{[Latitude, TenureMonths, MonthlyCharges, TotalCharges, CLTV, Gender, SeniorCitizen, Partner, Dependents, PhoneService, PaperlessBilling, MultipleLines, InternetService, OnlineSecurity, OnlineBackup, DeviceProtection, TechSupport, StreamingTV, StreamingMovies, Contract, PaymentMethod]}	
&
0.633 \\
\midrule
3
&
\parbox[t]{5cm}{[Latitude, Longitude, TenureMonths, MonthlyCharges, TotalCharges, CLTV, Gender, SeniorCitizen, Partner, Dependents, PhoneService, PaperlessBilling, MultipleLines, OnlineSecurity, DeviceProtection, TechSupport, StreamingTV, StreamingMovies, Contract, PaymentMethod]}	
&
0.638
&
\parbox[t]{5cm}{[Latitude, TenureMonths, MonthlyCharges, TotalCharges, CLTV, Gender, SeniorCitizen, Dependents, PhoneService, PaperlessBilling, MultipleLines, InternetService, OnlineSecurity, OnlineBackup, DeviceProtection, TechSupport, StreamingTV, StreamingMovies, Contract, PaymentMethod]}	
&
0.634 \\
\midrule

4
&
\parbox[t]{5cm}{[Latitude, Longitude, TenureMonths, MonthlyCharges, CLTV, Gender, SeniorCitizen, Partner, Dependents, PhoneService, PaperlessBilling, MultipleLines, OnlineSecurity, DeviceProtection, TechSupport, StreamingTV, StreamingMovies, Contract, PaymentMethod]}	
&
0.639
&
\parbox[t]{5cm}{[TenureMonths, MonthlyCharges, TotalCharges, CLTV, Gender, SeniorCitizen, Dependents, PhoneService, PaperlessBilling, MultipleLines, InternetService, OnlineSecurity, OnlineBackup, DeviceProtection, TechSupport, StreamingTV, StreamingMovies, Contract, PaymentMethod]}	
&
0.631 \\
\midrule


5
&
\parbox[t]{5cm}{[Latitude, Longitude, TenureMonths, MonthlyCharges, CLTV, Gender, SeniorCitizen, Partner, Dependents, PhoneService, PaperlessBilling, OnlineSecurity, DeviceProtection, TechSupport, StreamingTV, StreamingMovies, Contract, PaymentMethod]}	
&
0.639
&
\parbox[t]{5cm}{[TenureMonths, TotalCharges, CLTV, Gender, SeniorCitizen, Dependents, PhoneService, PaperlessBilling, MultipleLines, InternetService, OnlineSecurity, OnlineBackup, DeviceProtection, TechSupport, StreamingTV, StreamingMovies, Contract, PaymentMethod]}
&
0.630 \\
\midrule

6
&
\parbox[t]{5cm}{[Latitude, Longitude, TenureMonths, MonthlyCharges, CLTV, Gender, SeniorCitizen, Partner, Dependents, PhoneService, PaperlessBilling, OnlineSecurity, DeviceProtection, TechSupport, StreamingMovies, Contract, PaymentMethod]}
&
0.638
&
\parbox[t]{5cm}{[TenureMonths, TotalCharges, CLTV, Gender, SeniorCitizen, Dependents, PhoneService, PaperlessBilling, MultipleLines, InternetService, OnlineSecurity, DeviceProtection, TechSupport, StreamingTV, StreamingMovies, Contract, PaymentMethod]}
&
0.638 \\
\midrule

7
&
\parbox[t]{5cm}{\textcolor{darkgreen}{[Latitude, TenureMonths, MonthlyCharges, CLTV, Gender, SeniorCitizen, Partner, Dependents, PhoneService, PaperlessBilling, OnlineSecurity, DeviceProtection, TechSupport, StreamingMovies, Contract, PaymentMethod]}}
&
0.639
&
\parbox[t]{5cm}{\textcolor{darkgreen}{[TenureMonths, TotalCharges, CLTV, Gender, SeniorCitizen, Dependents, PhoneService, PaperlessBilling, MultipleLines, InternetService, OnlineSecurity, DeviceProtection, TechSupport, StreamingMovies, Contract, PaymentMethod]}}
& 
0.637 \\
\midrule

8
&
\parbox[t]{5cm}{[Latitude, TenureMonths, MonthlyCharges, CLTV, Gender, SeniorCitizen, Partner, Dependents, PhoneService, PaperlessBilling, OnlineSecurity, DeviceProtection, TechSupport, Contract, PaymentMethod]}
&
0.640
&
\parbox[t]{5cm}{[TenureMonths, TotalCharges, CLTV, Gender, SeniorCitizen, Dependents, PhoneService, PaperlessBilling, InternetService, OnlineSecurity, DeviceProtection, TechSupport, StreamingMovies, Contract, PaymentMethod]}
&
0.636 \\
\midrule

9
&
\parbox[t]{5cm}{[TenureMonths, MonthlyCharges, CLTV, Gender, SeniorCitizen, Partner, Dependents, PhoneService, PaperlessBilling, OnlineSecurity, DeviceProtection, TechSupport, Contract, PaymentMethod]}
&
0.639
&
\parbox[t]{5cm}{[TenureMonths, TotalCharges, CLTV, SeniorCitizen, Dependents, PhoneService, PaperlessBilling, InternetService, OnlineSecurity, DeviceProtection, TechSupport, StreamingMovies, Contract, PaymentMethod]}	
&
0.633 \\
\midrule


10
&
\parbox[t]{5cm}{[TenureMonths, MonthlyCharges, CLTV, Gender, Partner, Dependents, PhoneService, PaperlessBilling, OnlineSecurity, DeviceProtection, TechSupport, Contract, PaymentMethod]}
&
0.641
&
\parbox[t]{5cm}{[TenureMonths, TotalCharges, CLTV, SeniorCitizen, Dependents, PhoneService, PaperlessBilling, InternetService, OnlineSecurity, DeviceProtection, TechSupport, Contract, PaymentMethod]}
&
0.636 \\
\midrule

11
&
\parbox[t]{5cm}{[TenureMonths, MonthlyCharges, CLTV, Gender, Partner, Dependents, PhoneService, PaperlessBilling, DeviceProtection, TechSupport, Contract, PaymentMethod]}
&
0.642
&
\parbox[t]{5cm}{[TenureMonths, TotalCharges, CLTV, SeniorCitizen, Dependents, PhoneService, PaperlessBilling, OnlineSecurity, DeviceProtection, TechSupport, Contract, PaymentMethod]}
&
0.639 \\
\midrule

12
& 
\parbox[t]{5cm}{\textbf{[TenureMonths, MonthlyCharges, CLTV, Partner, Dependents, PhoneService, PaperlessBilling, DeviceProtection, TechSupport, Contract, PaymentMethod]}}
&
\textbf{0.644}
&
\parbox[t]{5cm}{\textbf{[TenureMonths, TotalCharges, CLTV, Dependents, PhoneService, PaperlessBilling, OnlineSecurity, DeviceProtection, TechSupport, Contract, PaymentMethod]}}
&
\textbf{0.640} \\
\midrule

13
&
\parbox[t]{5cm}{[TenureMonths, MonthlyCharges, CLTV, Partner, Dependents, PhoneService, PaperlessBilling, DeviceProtection, TechSupport, Contract]}
&
0.640
&
\parbox[t]{5cm}{[TenureMonths, TotalCharges, Dependents, PhoneService, PaperlessBilling, OnlineSecurity, DeviceProtection, TechSupport, Contract, PaymentMethod]}
&
0.637 \\
\midrule

14
&
\parbox[t]{5cm}{[TenureMonths, MonthlyCharges, CLTV, Partner, Dependents, PhoneService, DeviceProtection, TechSupport, Contract]}
&
0.641
&
\parbox[t]{5cm}{\textbf{[TenureMonths, TotalCharges, Dependents, PhoneService, OnlineSecurity, DeviceProtection, TechSupport, Contract, PaymentMethod]}}
&
\textbf{0.640} \\
\midrule

15
&
\parbox[t]{5cm}{[TenureMonths, MonthlyCharges, CLTV, Partner, Dependents, PhoneService, DeviceProtection, TechSupport]}
&
0.635
&
\parbox[t]{5cm}{[TenureMonths, TotalCharges, Dependents, PhoneService, OnlineSecurity, TechSupport, Contract, PaymentMethod]}
&
0.635 \\
\midrule

16
&
\parbox[t]{5cm}{[TenureMonths, MonthlyCharges, Partner, Dependents, PhoneService, DeviceProtection, TechSupport]}
&
0.633
&
\parbox[t]{5cm}{[TenureMonths, TotalCharges, Dependents, PhoneService, TechSupport, Contract, PaymentMethod]}
&
0.629 \\
\midrule

17
&
\parbox[t]{5cm}{[TenureMonths, MonthlyCharges, Partner, Dependents, PhoneService, TechSupport]}
&
0.624
&
\parbox[t]{5cm}{[TenureMonths, TotalCharges, Dependents, PhoneService, Contract, PaymentMethod]}
&
0.638 \\
\midrule

18
&
\parbox[t]{5cm}{[TenureMonths, MonthlyCharges, Partner, Dependents, TechSupport]}
&
0.628
&
\parbox[t]{5cm}{[TenureMonths, TotalCharges, Dependents, Contract, PaymentMethod]}
&
0.635 \\
\midrule

19
&
\parbox[t]{5cm}{[TenureMonths, MonthlyCharges, Dependents, TechSupport]}
&
0.626
&
\parbox[t]{5cm}{[TenureMonths, TotalCharges, Contract, PaymentMethod]}
&
0.629 \\
\midrule

20
&
\parbox[t]{5cm}{[TenureMonths, MonthlyCharges, Dependents]}
&
0.612
&
\parbox[t]{5cm}{[TenureMonths, TotalCharges, Contract]}
&
0.605 \\
\midrule

21
&
\parbox[t]{5cm}{[TenureMonths, MonthlyCharges]}
&
0.594
&
\parbox[t]{5cm}{[TotalCharges, Contract]}
&
0.582 \\
\midrule

22
&
\parbox[t]{5cm}{[TenureMonths]}
&
0.548
&
\parbox[t]{5cm}{[Contract]}
&
0.582 \\
\bottomrule
\end{longtable}
\end{centering}

\section{Experiment Results: MLP vs. RFC}
\label{sec:appendix_mlp_rfc}
\begin{figure}[H] 
  \centering
  \includegraphics[width=\textwidth]{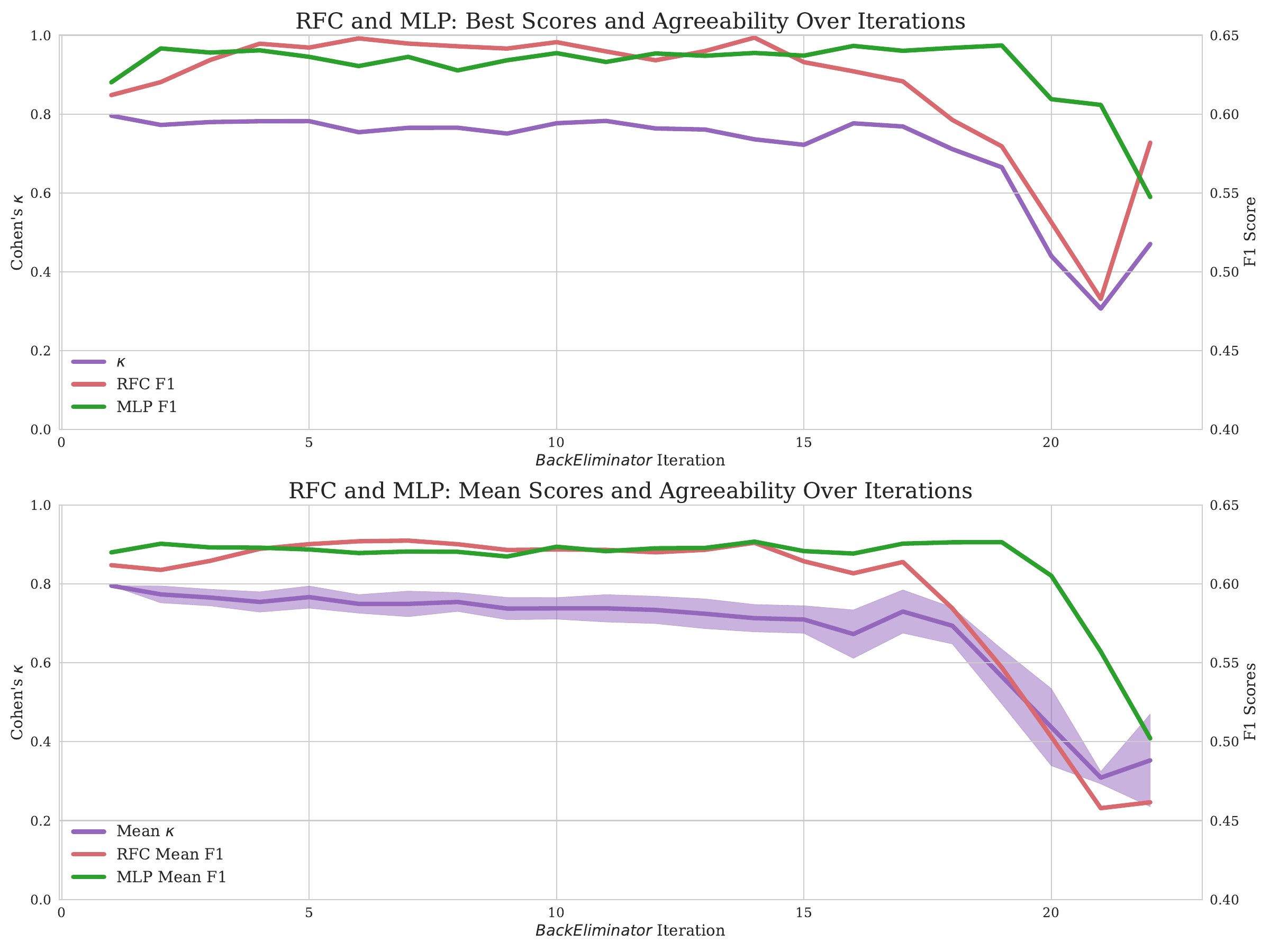}
  \caption{Best and mean \(F1\) (right-hand-side \(y\)-axis), \(\kappa\) (left-hand-side \(y\)-axis) scores from the MLP vs. RFC trial.}
  \label{fig:rfc_mlp}
\end{figure}

\begin{centering}    
\begin{longtable}{@{}c|p{6cm}|c|p{6cm}|c@{}}
\caption{Included features and performance of each model in MLP vs. RFC Trial. Bold typeface means best feature sets for \(F1\), dark green for best \(\kappa\).}
\label{tab:rfc_mlp_results} \\
\toprule
\multirow{2}{*}{\rotatebox[origin=c]{90}{Iter.}} & \multicolumn{2}{c|}{\textbf{MLP}} & \multicolumn{2}{c}{\textbf{RFC}}  \\

\cline{2-5}

&
\multicolumn{1}{c|}{Features} & \multicolumn{1}{c|}{\(F1\)} & \multicolumn{1}{c|}{Features} & \multicolumn{1}{c}{\(F1\)} \\

\endfirsthead

\multicolumn{5}{c}{Table \thetable\ continued from previous page} \\

\toprule
\multirow{2}{*}{\rotatebox[origin=c]{90}{Iter.}} & \multicolumn{2}{c|}{\textbf{MLP}} & \multicolumn{2}{c}{\textbf{RFC}}  \\

\cline{2-5}

&
\multicolumn{1}{c|}{Features} & \multicolumn{1}{c|}{\(F1\)} & \multicolumn{1}{c|}{Features} & \multicolumn{1}{c}{\(F1\)} \\
\midrule

\endhead
\bottomrule

1	
&
\parbox[t]{5cm}{\textcolor{darkgreen}{[Latitude, Longitude, TenureMonths, MonthlyCharges, TotalCharges, CLTV, Gender, SeniorCitizen, Partner, Dependents, PhoneService, PaperlessBilling, MultipleLines, InternetService, OnlineSecurity, OnlineBackup, DeviceProtection, TechSupport, StreamingTV, StreamingMovies, Contract, PaymentMethod]}}
&
0.620
&
\parbox[t]{5cm}{\textcolor{darkgreen}{[Latitude, Longitude, TenureMonths, MonthlyCharges, TotalCharges, CLTV, Gender, SeniorCitizen, Partner, Dependents, PhoneService, PaperlessBilling, MultipleLines, InternetService, OnlineSecurity, OnlineBackup, DeviceProtection, TechSupport, StreamingTV, StreamingMovies, Contract, PaymentMethod]}}
&
0.612 \\
\midrule
2
&
\parbox[t]{5cm}{[Latitude, Longitude, TenureMonths, MonthlyCharges, TotalCharges, CLTV, SeniorCitizen, Partner, Dependents, PhoneService, PaperlessBilling, MultipleLines, InternetService, OnlineSecurity, OnlineBackup, DeviceProtection, TechSupport, StreamingTV, StreamingMovies, Contract, PaymentMethod]}
&
0.642
&
\parbox[t]{5cm}{[Latitude, TenureMonths, MonthlyCharges, TotalCharges, CLTV, Gender, SeniorCitizen, Partner, Dependents, PhoneService, PaperlessBilling, MultipleLines, InternetService, OnlineSecurity, OnlineBackup, DeviceProtection, TechSupport, StreamingTV, StreamingMovies, Contract, PaymentMethod]}
&
0.620 \\
\midrule
3
&
\parbox[t]{5cm}{[Latitude, Longitude, TenureMonths, MonthlyCharges, TotalCharges, CLTV, SeniorCitizen, Partner, Dependents, PhoneService, MultipleLines, InternetService, OnlineSecurity, OnlineBackup, DeviceProtection, TechSupport, StreamingTV, StreamingMovies, Contract, PaymentMethod]}
&
0.639
&
\parbox[t]{5cm}{[Latitude, TenureMonths, MonthlyCharges, TotalCharges, CLTV, Gender, SeniorCitizen, Partner, Dependents, PhoneService, PaperlessBilling, MultipleLines, InternetService, OnlineBackup, DeviceProtection, TechSupport, StreamingTV, StreamingMovies, Contract, PaymentMethod]}
&
0.634 \\
\midrule
4
&
\parbox[t]{5cm}{[Latitude, Longitude, TenureMonths, MonthlyCharges, TotalCharges, CLTV, SeniorCitizen, Partner, Dependents, PhoneService, MultipleLines, InternetService, OnlineBackup, DeviceProtection, TechSupport, StreamingTV, StreamingMovies, Contract, PaymentMethod]}
&
0.640
&
\parbox[t]{5cm}{[Latitude, TenureMonths, MonthlyCharges, TotalCharges, CLTV, SeniorCitizen, Partner, Dependents, PhoneService, PaperlessBilling, MultipleLines, InternetService, OnlineBackup, DeviceProtection, TechSupport, StreamingTV, StreamingMovies, Contract, PaymentMethod]}
&
0.645 \\
\midrule
5
&
\parbox[t]{5cm}{[Latitude, Longitude, TenureMonths, MonthlyCharges, TotalCharges, CLTV, SeniorCitizen, Partner, Dependents, PhoneService, MultipleLines, OnlineBackup, DeviceProtection, TechSupport, StreamingTV, StreamingMovies, Contract, PaymentMethod]}
&
0.636
&
\parbox[t]{5cm}{[Latitude, TenureMonths, MonthlyCharges, TotalCharges, CLTV, SeniorCitizen, Partner, Dependents, PhoneService, PaperlessBilling, MultipleLines, InternetService, OnlineBackup, TechSupport, StreamingTV, StreamingMovies, Contract, PaymentMethod]}
&
0.642 \\
\midrule
6
&
\parbox[t]{5cm}{[Latitude, Longitude, TenureMonths, MonthlyCharges, TotalCharges, CLTV, SeniorCitizen, Partner, Dependents, PhoneService, MultipleLines, OnlineBackup, DeviceProtections, TechSupport, StreamingTV, Contract, PaymentMethod]}
&
0.630
&
\parbox[t]{5cm}{\textbf{[Latitude, TenureMonths, MonthlyCharges, TotalCharges, CLTV, Partner, Dependents, PhoneService, PaperlessBilling, MultipleLines, InternetService, OnlineBackup, TechSupport, StreamingTV, StreamingMovies, Contract, PaymentMethod]}}
&
\textbf{0.648} \\
\midrule
7
&
\parbox[t]{5cm}{[Latitude, Longitude, TenureMonths, MonthlyCharges, TotalCharges, SeniorCitizen, Partner, Dependents, PhoneService, MultipleLines, OnlineBackup, DeviceProtection, TechSupport, StreamingTV, Contract, PaymentMethod]}
&
0.636
&
\parbox[t]{5cm}{[Latitude, TenureMonths, MonthlyCharges, TotalCharges, CLTV, Partner, Dependents, PaperlessBilling, MultipleLines, InternetService, OnlineBackup, TechSupport, StreamingTV, StreamingMovies, Contract, PaymentMethod]}
&
0.645 \\
\midrule
8
&
\parbox[t]{5cm}{[Latitude, Longitude, TenureMonths, MonthlyCharges, SeniorCitizen, Partner, Dependents, PhoneService, MultipleLines, OnlineBackup, DeviceProtection, TechSupport, StreamingTV, Contract, PaymentMethod]}
&
0.628
&
\parbox[t]{5cm}{[Latitude, TenureMonths, MonthlyCharges, TotalCharges, CLTV, Dependents, PaperlessBilling, MultipleLines, InternetService, OnlineBackup, TechSupport, StreamingTV, StreamingMovies, Contract, PaymentMethod]}
&
0.643 \\
\midrule
9
&
\parbox[t]{5cm}{[Longitude, TenureMonths, MonthlyCharges, SeniorCitizen, Partner, Dependents, PhoneService, MultipleLines, OnlineBackup, DeviceProtection, TechSupport, StreamingTV, Contract, PaymentMethod]}
&
0.634
&
\parbox[t]{5cm}{[Latitude, TenureMonths, MonthlyCharges, TotalCharges, CLTV, Dependents, PaperlessBilling, MultipleLines, InternetService, OnlineBackup, TechSupport, StreamingMovies, Contract, PaymentMethod]}
&
0.641 \\
\midrule
10
&
\parbox[t]{5cm}{[Longitude, TenureMonths, MonthlyCharges, Partner, Dependents, PhoneService, MultipleLines, OnlineBackup, DeviceProtection, TechSupport, StreamingTV, Contract, PaymentMethod]}
&
0.639
&
\parbox[t]{5cm}{[Latitude, TenureMonths, MonthlyCharges, TotalCharges, CLTV, Dependents, PaperlessBilling, MultipleLines, InternetService, OnlineBackup, TechSupport, Contract, PaymentMethod]}
&
0.646 \\
\midrule
11
&
\parbox[t]{5cm}{[Longitude, TenureMonths, MonthlyCharges, Partner, Dependents, PhoneService, MultipleLines, OnlineBackup, TechSupport, StreamingTV, Contract, PaymentMethod]}
&
0.633
&
\parbox[t]{5cm}{[Latitude, TenureMonths, MonthlyCharges, TotalCharges, CLTV, Dependents, MultipleLines, InternetService, OnlineBackup, TechSupport, Contract, PaymentMethod]}
&
0.640 \\
\midrule
12
&
\parbox[t]{5cm}{[TenureMonths, MonthlyCharges, Partner, Dependents, PhoneService, MultipleLines, OnlineBackup, TechSupport, StreamingTV, Contract, PaymentMethod]}
&
0.638
&
\parbox[t]{5cm}{[Latitude, TenureMonths, MonthlyCharges, TotalCharges, CLTV, Dependents, MultipleLines, InternetService, TechSupport, Contract, PaymentMethod]}
&
0.634 \\
\midrule
13
&
\parbox[t]{5cm}{[TenureMonths, MonthlyCharges, Partner, Dependents, PhoneService, OnlineBackup, TechSupport, StreamingTV, Contract, PaymentMethod]}
&
0.637
&
\parbox[t]{5cm}{[Latitude, TenureMonths, MonthlyCharges, TotalCharges, CLTV, Dependents, InternetService, TechSupport, Contract, PaymentMethod]}
&
0.640 \\
\midrule
14
&
\parbox[t]{5cm}{[TenureMonths, MonthlyCharges, Dependents, PhoneService, OnlineBackup, TechSupport, StreamingTV, Contract, PaymentMethod]}
&
0.639
&
\parbox[t]{5cm}{\textbf{[Latitude, TenureMonths, MonthlyCharges, CLTV, Dependents, InternetService, TechSupport, Contract, PaymentMethod]}}
&
\textbf{0.648} \\
\midrule
15
&
\parbox[t]{5cm}{[TenureMonths, MonthlyCharges, Dependents, PhoneService, OnlineBackup TechSupport, StreamingTV, Contract]}
&
0.637
&
\parbox[t]{5cm}{[Latitude, TenureMonths, MonthlyCharges, CLTV, Dependents, TechSupport, Contract, PaymentMethod]}
&
0.633 \\
\midrule
16
&
\parbox[t]{5cm}{\textbf{[TenureMonths, MonthlyCharges, Dependents, PhoneService, TechSupport, StreamingTV, Contract]}}
&
\textbf{0.643}
&
\parbox[t]{5cm}{[Latitude, TenureMonths, MonthlyCharges, CLTV, Dependents, TechSupport, Contract]}
&
0.627 \\
\midrule
17
&
\parbox[t]{5cm}{[TenureMonths, MonthlyCharges, Dependents, TechSupport, StreamingTV, Contract]}
&
0.640
&
\parbox[t]{5cm}{[TenureMonths, MonthlyCharges, CLTV, Dependents, TechSupport, Contract]}
&
0.621 \\
\midrule
18
&
\parbox[t]{5cm}{[TenureMonths, MonthlyCharges, Dependents, TechSupport, Contract]}
&
0.642
&
\parbox[t]{5cm}{[MonthlyCharges, CLTV, Dependents, TechSupport, Contract]}
&
0.596 \\
\midrule
19
&
\parbox[t]{5cm}{\textbf{[TenureMonths, MonthlyCharges, Dependents, Contract]}}
&
\textbf{0.643}
&
\parbox[t]{5cm}{[MonthlyCharges, CLTV, Dependents, Contract]}
&
0.579 \\
\midrule
20
&
\parbox[t]{5cm}{[TenureMonths, MonthlyCharges, Dependents]}
&
0.609
&
\parbox[t]{5cm}{[MonthlyCharges, CLTV, Contract]}
&
0.531 \\
\midrule
21
&
[TenureMonths, MonthlyCharges]
&
0.606
&
[MonthlyCharges, Contract]
&
0.483 \\
\midrule
22
&
\parbox[t]{5cm}{[TenureMonths]}
&
0.547
&
\parbox[t]{5cm}{[Contract]}
&
0.582 \\
\bottomrule
\end{longtable}
\end{centering}

\section{Experiment Results: RFC vs. Logit}
\label{sec:appendix_logit_rfc}
\begin{figure}[H] 
  \centering
  \includegraphics[width=\textwidth]{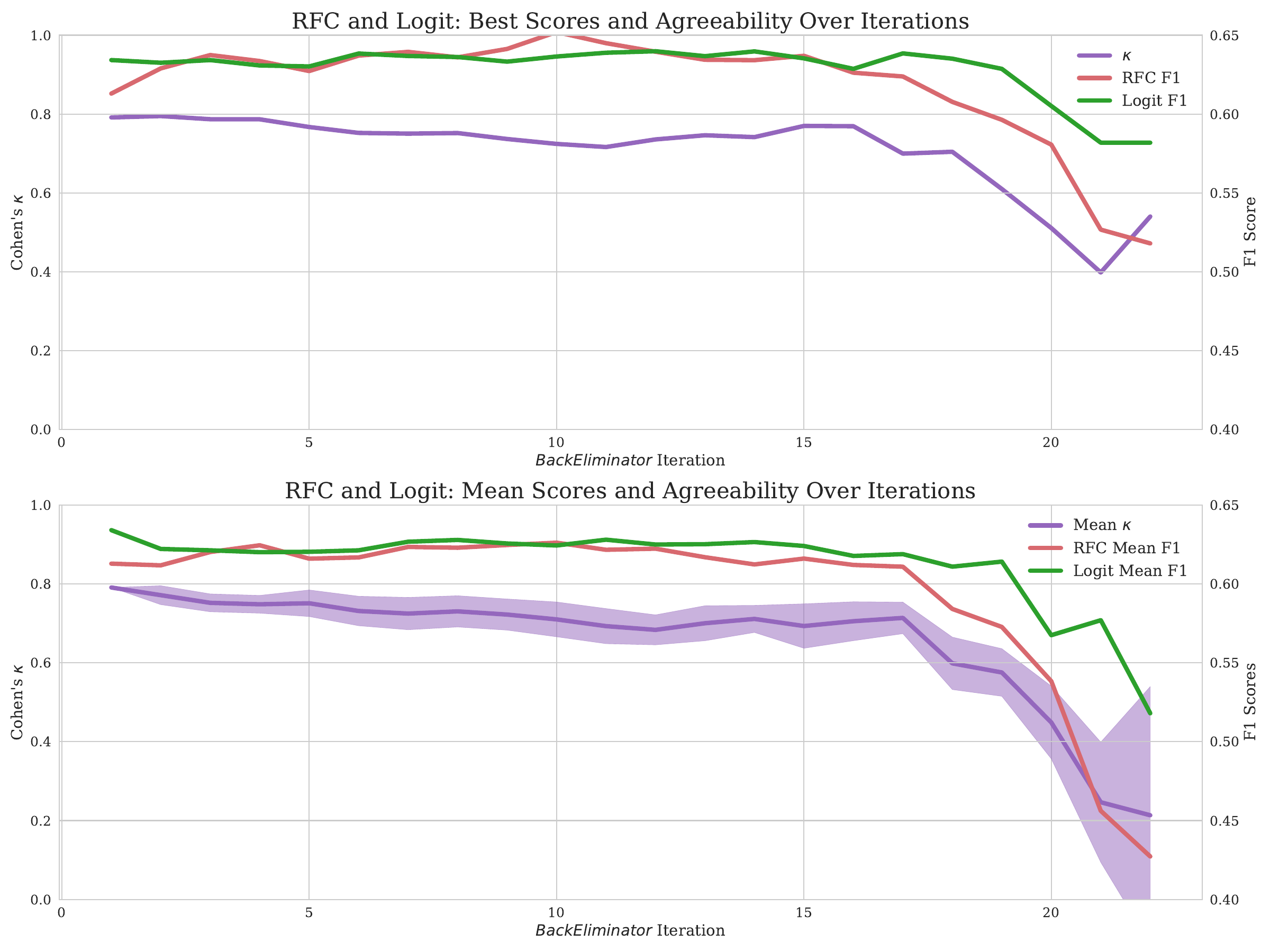}
  \caption{Best and mean \(F1\) (right-hand-side \(y\)-axis), \(\kappa\) (left-hand-side \(y\)-axis) scores from the Logit vs. RFC trial.}
  \label{fig:rfc_logit}
\end{figure}

\begin{centering}    
\begin{longtable}{@{}c|p{6cm}|c|p{6cm}|c@{}}
\caption{Included features and performance of each model in RFC vs. Logit Trial. Bold typeface means best feature sets for \(F1\), dark green for best \(\kappa\).}
\label{tab:rfc_logit_results} \\
\toprule
\multirow{2}{*}{\rotatebox[origin=c]{90}{Iter.}} & \multicolumn{2}{c|}{\textbf{RFC}} & \multicolumn{2}{c}{\textbf{Logit}}  \\

\cline{2-5}

&
\multicolumn{1}{c|}{Features} & \multicolumn{1}{c|}{\(F1\)} & \multicolumn{1}{c|}{Features} & \multicolumn{1}{c}{\(F1\)} \\

\endfirsthead

\multicolumn{5}{c}{Table \thetable\ continued from previous page} \\

\toprule
\multirow{2}{*}{\rotatebox[origin=c]{90}{Iter.}} & \multicolumn{2}{c|}{\textbf{RFC}} & \multicolumn{2}{c}{\textbf{Logit}}  \\

\cline{2-5}

&
\multicolumn{1}{c|}{Features} & \multicolumn{1}{c|}{\(F1\)} & \multicolumn{1}{c|}{Features} & \multicolumn{1}{c}{\(F1\)} \\
\midrule

\endhead
\bottomrule
1
&
\parbox[t]{5cm}{[Latitude, Longitude, TenureMonths, MonthlyCharges, TotalCharges, CLTV, Gender, SeniorCitizen, Partner, Dependents, PhoneService, PaperlessBilling, MultipleLines, InternetService, OnlineSecurity, OnlineBackup, DeviceProtection, TechSupport, StreamingTV, StreamingMovies, Contract, PaymentMethod]}
&
0.613
&
\parbox[t]{5cm}{[Latitude, Longitude, TenureMonths, MonthlyCharges, TotalCharges, CLTV, Gender, SeniorCitizen, Partner, Dependents, PhoneService, PaperlessBilling, MultipleLines, InternetService, OnlineSecurity, OnlineBackup, DeviceProtection, TechSupport, StreamingTV, StreamingMovies, Contract, PaymentMethod]}
&
0.634 \\
\midrule
2
&
\parbox[t]{5cm}{\textcolor{darkgreen}{[Latitude, Longitude, TenureMonths, MonthlyCharges, TotalCharges, CLTV, Gender, SeniorCitizen, Partner, Dependents, PhoneService, PaperlessBilling, MultipleLines, InternetService, OnlineBackup, DeviceProtection, TechSupport, StreamingTV, StreamingMovies, Contract, PaymentMethod]}}
&
0.629
&
\parbox[t]{5cm}{\textcolor{darkgreen}{[Latitude, TenureMonths, MonthlyCharges, TotalCharges, CLTV, Gender, SeniorCitizen, Partner, Dependents, PhoneService, PaperlessBilling, MultipleLines, InternetService, OnlineSecurity, OnlineBackup, DeviceProtection, TechSupport, StreamingTV, StreamingMovies, Contract, PaymentMethod]}}
&
0.633 \\
\midrule
3
&
\parbox[t]{5cm}{[Latitude, Longitude, TenureMonths, MonthlyCharges, TotalCharges, CLTV, Gender, Partner, Dependents, PhoneService, PaperlessBilling, MultipleLines, InternetServiceo, OnlineBackup, DeviceProtection, TechSupport, StreamingTV, StreamingMovies, Contract, PaymentMethod]}
&
0.637
&
\parbox[t]{5cm}{[Latitude, TenureMonths, MonthlyCharges, TotalCharges, CLTV, Gender, SeniorCitizen, Dependents, PhoneService, PaperlessBilling, MultipleLines, InternetService, OnlineSecurity, OnlineBackup, DeviceProtection, TechSupport, StreamingTV, StreamingMovies, Contract, PaymentMethod]}
&
0.634 \\
\midrule
4
&
\parbox[t]{5cm}{[Latitude, Longitude, TenureMonths, MonthlyCharges, CLTV, Gender, Partner, Dependents, PhoneService, PaperlessBilling, MultipleLines, InternetService, OnlineBackup, DeviceProtection, TechSupport, StreamingTV, StreamingMovies, Contract, PaymentMethod]}
&
0.634
&
\parbox[t]{5cm}{[TenureMonths, MonthlyCharges, TotalCharges, CLTV, Gender, SeniorCitizen, Dependents, PhoneService, PaperlessBilling, MultipleLines, InternetService, OnlineSecurity, OnlineBackup, DeviceProtection, TechSupport, StreamingTV, StreamingMovies, Contract, PaymentMethod]}
&
0.631 \\
\midrule
5
&
\parbox[t]{5cm}{[Latitude, Longitude, TenureMonths, MonthlyCharges, CLTV, Gender, Partner, Dependents, PhoneService, PaperlessBilling, MultipleLines, InternetService, OnlineBackup, DeviceProtection, TechSupport, StreamingTV, Contract, PaymentMethod]}
&
0.627
&
\parbox[t]{5cm}{[TenureMonths, TotalCharges, CLTV, Gender, SeniorCitizen, Dependents, PhoneService, PaperlessBilling, MultipleLines, InternetService, OnlineSecurity, OnlineBackup, DeviceProtection, TechSupport, StreamingTV, StreamingMovies, Contract, PaymentMethod]}
&
0.630 \\
\midrule
6
&
\parbox[t]{5cm}{[Latitude, Longitude, TenureMonths, MonthlyCharges, CLTV, Partner, Dependents, PhoneService, PaperlessBilling, MultipleLines, InternetService, OnlineBackup, DeviceProtection, TechSupport, StreamingTV, Contract, PaymentMethod]}
&
0.637
&
\parbox[t]{5cm}{[TenureMonths, TotalCharges, CLTV, Gender, SeniorCitizen, Dependents, PhoneService, PaperlessBilling, MultipleLines, InternetService, OnlineSecurity, DeviceProtection, TechSupport, StreamingTV, StreamingMovies, Contract, PaymentMethod]}
&
0.638 \\
\midrule
7
&
\parbox[t]{5cm}{[Latitude, Longitude, TenureMonths, MonthlyCharges, CLTV, Partner, Dependents, PhoneService, PaperlessBilling, MultipleLines, InternetServiceo, OnlineBackup, DeviceProtection, TechSupport, StreamingTV, Contract]}
&
0.639
&
\parbox[t]{5cm}{[TenureMonths, TotalCharges, CLTV, Gender, SeniorCitizen, Dependents, PhoneService, PaperlessBilling, MultipleLines, InternetService, OnlineSecurity, DeviceProtection, TechSupport, StreamingMovies, Contract, PaymentMethod]}
&
0.637 \\
\midrule
8
&
\parbox[t]{5cm}{[Latitude, Longitude, TenureMonths, MonthlyCharges, CLTV, Partner, Dependents, PhoneService, PaperlessBilling, MultipleLines, InternetService, DeviceProtection, TechSupport, StreamingTV, Contract]}
&
0.636
&
\parbox[t]{5cm}{[TenureMonths, TotalCharges, CLTV, Gender, SeniorCitizen, Dependents, PhoneService, PaperlessBilling, InternetService, OnlineSecurity, DeviceProtection, TechSupport, StreamingMovies, Contract, PaymentMethod]}
&
0.636 \\
\midrule

9
&
\parbox[t]{5cm}{[Latitude, TenureMonths, MonthlyCharges, CLTV, Partner, Dependents, PhoneService, PaperlessBilling, MultipleLines, InternetService, DeviceProtection, TechSupport, StreamingTV, Contract]}
&
0.641
&
\parbox[t]{5cm}{[TenureMonths, TotalCharges, CLTV, SeniorCitizen, Dependents, PhoneService, PaperlessBilling, InternetService, OnlineSecurity, DeviceProtection, TechSupport, StreamingMovies, Contract, PaymentMethod]}
&
0.633 \\
\midrule

10
&
\parbox[t]{5cm}{\textbf{[Latitude, TenureMonths, MonthlyCharges, CLTV, Partner, Dependents, PhoneService, PaperlessBilling, MultipleLines, InternetService, TechSupport, StreamingTV, Contract]}}
&
\textbf{0.652}
&
\parbox[t]{5cm}{[TenureMonths, TotalCharges, CLTV, SeniorCitizen, Dependents, PhoneService, PaperlessBilling, InternetService, OnlineSecurity, DeviceProtection, TechSupport, Contract, PaymentMethod]}
&
0.636 \\
\midrule

11
&
\parbox[t]{5cm}{[Latitude, TenureMonths, MonthlyCharges, CLTV, Dependents, PhoneService, PaperlessBilling, MultipleLines, InternetService, TechSupport, StreamingTV, Contract]}
&
0.645
&
\parbox[t]{5cm}{[TenureMonths, TotalCharges, CLTV, SeniorCitizen, Dependents, PhoneService, PaperlessBilling, OnlineSecurity, DeviceProtection, TechSupport, Contract, PaymentMethod]}
&
0.639 \\
\midrule

12
&
\parbox[t]{5cm}{[Latitude, TenureMonths, MonthlyCharges, CLTV, Dependents, PhoneService, PaperlessBilling, MultipleLines, TechSupport, StreamingTV, Contract]}
&
0.640
&
\parbox[t]{5cm}{\textbf{[TenureMonths, TotalCharges, CLTV, Dependents, PhoneService, PaperlessBilling, OnlineSecurity, DeviceProtection, TechSupport, Contract, PaymentMethod]}}
&
\textbf{0.640} \\
\midrule

13
&
\parbox[t]{5cm}{[Latitude, TenureMonths, MonthlyCharges, CLTV, Dependents, PhoneService, PaperlessBilling, TechSupport, StreamingTV, Contract]}
&
0.634
&
\parbox[t]{5cm}{[TenureMonths, TotalCharges, Dependents, PhoneService, PaperlessBilling, OnlineSecurity, DeviceProtection, TechSupport, Contract, PaymentMethod]}
&
0.637 \\
\midrule

14
&
\parbox[t]{5cm}{[Latitude, TenureMonths, MonthlyCharges, CLTV, Dependents, PhoneService, PaperlessBilling, TechSupport, Contract]}
&
0.634
&
\parbox[t]{5cm}{\textbf{[TenureMonths, TotalCharges, Dependents, PhoneService, OnlineSecurity, DeviceProtection, TechSupport, Contract, PaymentMethod]}}
&
\textbf{0.640} \\
\midrule

15
&
\parbox[t]{5cm}{[Latitude, TenureMonths, MonthlyCharges, CLTV, Dependents, PaperlessBilling, TechSupport, Contract]}
&
0.637
&
\parbox[t]{5cm}{[TenureMonths, TotalCharges, Dependents, PhoneService, OnlineSecurity, TechSupport, Contract, PaymentMethod]}
&
0.635 \\
\midrule

16
&
\parbox[t]{5cm}{[Latitude, TenureMonths, MonthlyCharges, CLTV, Dependents, TechSupport, Contract]}
&
0.626
&
\parbox[t]{5cm}{[TenureMonths, TotalCharges, Dependents, PhoneService, TechSupport, Contract, PaymentMethod]}
&
0.629 \\
\midrule

17
&
\parbox[t]{5cm}{[Latitude, TenureMonths, MonthlyCharges, Dependents, TechSupport, Contract]}
&
0.624
&
\parbox[t]{5cm}{[TenureMonths, TotalCharges, Dependents, PhoneService, Contract, PaymentMethod]}
&
0.638 \\
\midrule

18
&
\parbox[t]{5cm}{[Latitude, TenureMonths, MonthlyCharges, Dependents, Contract]}
&
0.608
&
\parbox[t]{5cm}{[TenureMonths, TotalCharges, Dependents, Contract, PaymentMethod]}
&
0.635 \\
\midrule

19
&
\parbox[t]{5cm}{[Latitude, TenureMonths, MonthlyCharges, Contract]}
&
0.596
&
\parbox[t]{5cm}{[TenureMonths, TotalCharges, Contract, PaymentMethod]}
&
0.629 \\
\midrule

20
&
\parbox[t]{5cm}{[Latitude, TenureMonths, MonthlyCharges]}
&
0.581
&
\parbox[t]{5cm}{[TenureMonths, TotalCharges, Contract]}
&
0.605 \\
\midrule

21
&
\parbox[t]{5cm}{[TenureMonths, MonthlyCharges]}
&
0.527
&
\parbox[t]{5cm}{[TotalCharges, Contract]}
&
0.582 \\
\midrule

22
&
\parbox[t]{5cm}{[TenureMonths]}
&
0.518
&
\parbox[t]{5cm}{[Contract]}
&
0.582 \\
\bottomrule
\end{longtable}
\end{centering}

\end{document}